\documentclass[10pt,twocolumn,letterpaper]{article}

\usepackage{wacv}

\usepackage{times}
\usepackage{epsfig}
\usepackage{graphicx}
\usepackage{amsmath}
\usepackage{amssymb}
\usepackage{booktabs}

\usepackage{amsfonts}       
\usepackage{adjustbox}
\usepackage{epsfig}
\usepackage{multirow}
\usepackage{makecell}
\usepackage{siunitx}
\usepackage{textcomp}
\usepackage{diagbox}
\usepackage{colortbl}
\usepackage{collcell}
\usepackage{bbm}
\usepackage{wrapfig}
\usepackage{resizegather}
\usepackage{float}
\usepackage[dvipsnames]{xcolor}         
\usepackage{comment}
\usepackage{arydshln} 

\makeatletter
\@namedef{ver@everyshi.sty}{}
\makeatother
\usepackage{tikz}

\def\method{MixOE}
\def\upcolor{LimeGreen}
\def\downcolor{OrangeRed}
\definecolor{coral}{rgb}{1.0, 0.5, 0.31}
\definecolor{high}{HTML}{B13230}
\definecolor{low}{HTML}{FEE8C3}

\def\eg{\textit{e.g.}}
\def\ie{\textit{i.e.}}
\def\etal{\textit{et al.}~}

\newcommand{\bird}{\textsl{Bird}}
\newcommand{\butterfly}{\textsl{Butterfly}}
\newcommand{\car}{\textsl{Car}}
\newcommand{\aircraft}{\textsl{Aircraft}}

\newcommand{\Din}{\mathcal{D}_{\text{in}}}

\newcommand{\Dio}{\mathcal{D}_{\text{out}}^{\text{virtual}}}

\def\wacvPaperID{732} 

\wacvalgorithmstrack   

\wacvfinalcopy 

\ifwacvfinal
\usepackage[breaklinks=true,bookmarks=false]{hyperref}
\else
\usepackage[pagebackref=true,breaklinks=true,colorlinks,bookmarks=false]{hyperref}
\fi

\begin{document}

\title{Mixture Outlier Exposure: Towards Out-of-Distribution Detection in Fine-grained Environments}

\author{Jingyang Zhang\textsuperscript{\textdagger}, Nathan Inkawhich\textsuperscript{*}, Randolph Linderman\textsuperscript{\textdagger}, Yiran Chen\textsuperscript{\textdagger}, Hai Li\textsuperscript{\textdagger}\\
\textsuperscript{\textdagger}Duke University, 
\textsuperscript{*}Air Force Research Laboratory\\
{\tt\small jingyang.zhang@duke.edu}
}

\maketitle
\thispagestyle{empty}

\begin{abstract}
   Many real-world scenarios in which DNN-based recognition systems are deployed have inherently fine-grained attributes (\eg, bird-species recognition, medical image classification).
    In addition to achieving reliable accuracy, a critical subtask for these models is to detect Out-of-distribution (OOD) inputs.
    Given the nature of the deployment environment, one may expect such OOD inputs to also be fine-grained w.r.t. the known classes (\eg, a novel bird species), which are thus extremely difficult to identify.
    Unfortunately, OOD detection in fine-grained scenarios remains largely underexplored.
    In this work, we aim to fill this gap by first carefully constructing four large-scale fine-grained test environments, in which existing methods are shown to have difficulties.
    Particularly, we find that even explicitly incorporating a diverse set of auxiliary outlier data during training does not provide sufficient coverage over the broad region where fine-grained OOD samples locate.
    We then propose Mixture Outlier Exposure (MixOE), which mixes ID data and training outliers to expand the coverage of different OOD granularities, and trains the model such that the prediction confidence linearly decays as the input transitions from ID to OOD.
    Extensive experiments and analyses demonstrate the effectiveness of MixOE for building up OOD detector in fine-grained environments. The code is available at \url{https://github.com/zjysteven/MixOE}.
\end{abstract}

\section{Introduction}
\label{sec:intro}

Real-world scenarios in which DNN recognition systems are deployed are often fine-grained in nature, wherein the data coming from such environments share a high level of semantic/visual similarity.
Examples include fine-grained visual classification \cite{zheng2017learning,hsu2019fine,yang2021re}, medical image classification \cite{li2014medical,fan2020microscopic}, and remote sensing applications \cite{zhu2017deep,xia2017aid,xia2018dota}.
In addition to achieving accurate classification, a critical problem for DNN models is to identify out-of-distribution (OOD) samples during inference time which do not belong to one of the DNN's known classes.
Such detection is crucial for building safe and reliable intelligent systems that operate in the open world.
However, we posit that OOD detection is particularly challenging in fine-grained scenarios because one may expect the OOD inputs to be highly granular w.r.t. the in-distribution (ID) data (\eg, novel bird species to a bird classifier), given the nature of the deployment environment. 
Such fine-grained OOD samples can make detection much difficult because they may use very similar feature sets to the ID data \cite{inkawhich2021advoe}.

\begin{figure*}[!t]
  \centering
  \includegraphics[width=.9\linewidth]{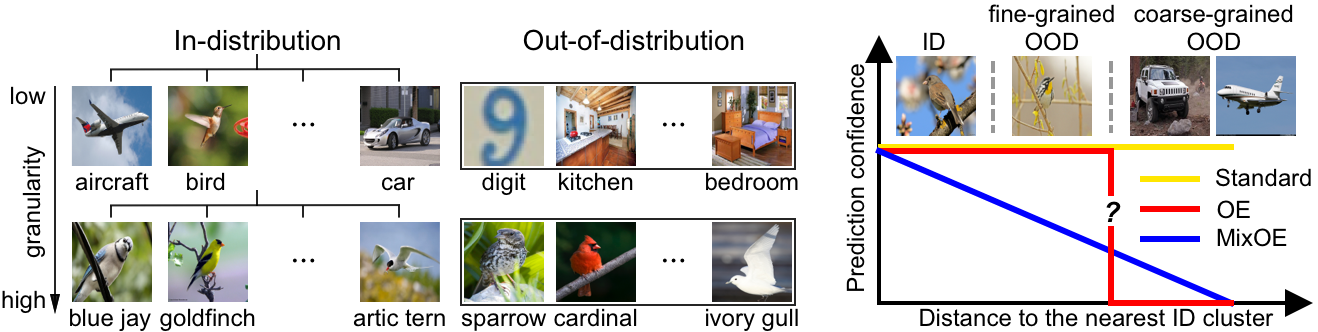}
  \caption{\textbf{Left:} A comparison of OOD detection in coarse- and fine-grained environments. Intuitively, fine-grained detection is much more challenging.
  \textbf{Right:} A \textit{conceptual} illustration of MixOE. A Standard model with no OOD considerations tends to be over-confident on OOD samples. OE is able to calibrate the prediction confidence on coarse-grained OOD, but the outputs on fine-grained OOD are not explicitly controlled (marked by ``?''). MixOE aims for a smooth decay of the confidence as the inputs transition from ID to OOD, and thus enables detection of both coarse/fine-grained OOD.}
  \label{fig:motivation}
\end{figure*}

Unfortunately, despite being inherently motivated by many real-world scenarios, \textit{OOD detection in fine-grained environments remains largely underexplored/underconsidered in current research.}
This in part has to do with (overly) simplistic \cite{ahmed2020detecting,tack2020csi} and coarse-grained benchmarks that are currently being used to evaluate OOD detection methods \cite{liang2018enhancing,lee2018simple,sastry2020detecting,liu2020energy,hsu2020generalized,winkens2020contrastive,techapanurak2021bridging} (\eg, CIFAR-10/100 v.s. SVHN/LSUN).
An illustrative comparison between OOD detection in fine- and coarse-grained environments is shown in Fig. \ref{fig:motivation}.
A few recent works \cite{perera2019deep,ahmed2020detecting,techapanurak2021practical} did consider fine-grained settings but did \textit{not} carefully/thoroughly investigate why and how they are difficult.
Besides, these works either operated at a rather limited scale, \ie, very few ID classes were considered \cite{ahmed2020detecting,techapanurak2021practical}, or made unrealistic practices such as training on a \textit{labeled} outlier dataset that overlaps with testing OOD data distribution \cite{perera2019deep}.

In this work, \textit{our goal is to fill the gap and present (to our knowledge) the first study that specifically targets OOD detection in fine-grained environments.}
We start by building four large-scale, fine-grained test environments to approximate real-world scenarios (Sec. \ref{sec:S3}). 
We find that several state-of-the-art OOD detection methods struggle to detect fine-grained novelties, which highlights the challenging nature of OOD detection in fine-grained settings.
Then, through analysis we identify that fine-grained OOD samples span a much broader region and are closer to the ID clusters in the DNN's feature space (Sec. \ref{sec:4.1}).
In addition, we find that including a large/diverse set of outlier data during training  \cite{hendrycks2018deep,chen2021atom,liu2020energy} does \textit{not} help cover the area where fine-grained OOD data locate.

Finally, we design a novel training algorithm named \textit{Mixture Outlier Exposure} (\method) to address the observed issue in fine-grained OOD detection (Sec. \ref{sec:4.2}).
Specifically, we propose to perform \textit{mixing operations} (in this work we adopt Mixup \cite{zhang2018mixup} or CutMix \cite{yun2019cutmix}) between ID data and outlier data to get ``virtual'' outlier samples which can be both near to and far away from the ID clusters.
The model is then trained such that the prediction confidence linearly decays as the input transitions from ID to OOD (see Fig. \ref{fig:motivation} for illustration).
As such, MixOE induces regularization over a larger OOD region and has clear implications for both coarse- and fine-grained OOD samples.
Experimental results on the four test benches show that a simple fine-tuning with MixOE can lead to consistently higher or competitive detection rates against both coarse- and fine-grained OOD data (Sec. \ref{sec:S5}).
We also conduct careful ablation study to further understand why and how MixOE works.
Our contributions are summarized as follows:
\begin{itemize}
    \item We construct four large-scale test environments for fine-grained OOD detection, which can be easily generated from existing public datasets and facilitate future studies;
    \item We propose MixOE, a novel OOD detection methodology that has effect across a spectrum of OOD granularities;
    \item We show that MixOE leads to notable improvements on all four benchmarks, in particular against fine-grained OOD where few current methods have any impacts.
\end{itemize}
\section{Related work}
\label{sec:S2}

\noindent \textbf{OOD detection approaches.}
Many popular works in OOD detection research use a pre-trained DNN classifier as a base model, and design an OOD scoring mechanism that leverages some signal from this model.
Several methods utilize the output space of the classifier, \eg, MSP \cite{hendrycks2018deep}, ODIN \cite{liang2018enhancing}, and Energy \cite{liu2020energy}, while other works such as Mahalanobis detector \cite{lee2018simple} and Gram Matrices \cite{sastry2020detecting} focus on the intermediate feature space of DNNs.
Recent works also start to explore the potential of gradient information \cite{huang2021importance}.

Another line of research modifies the base DNN's training phase to enable better detection.
Lee \etal \cite{lee2018training} proposed to synthesize OOD samples with a GAN \cite{radford2015unsupervised} and force the classifier to be less confident on the generated OOD data.
Hendrycks \etal \cite{hendrycks2018deep} later showed that a diverse and realistic outlier distribution is more proficient than synthetic samples at encouraging low confidence predictions on unseen OOD data.
More recently there comes the idea of VOS \cite{du2022towards}, which does not use real data but generates virtual outliers by sampling from class-conditional Gaussian distribution.
However, the assumption that ID data follow class-conditional Gaussian may not hold especially when the number of classes is large, and VOS has shown to even have trouble scaling to CIFAR-100 \cite{vos_opinion}.
To conclude, incorporating auxiliary outlier data in training is still one of the most effective methods to date \cite{hendrycks2018deep,chen2021atom,liu2020energy}.

Our work is most closely related to \cite{hendrycks2018deep} in that we too use \textit{unlabeled} auxiliary outlier data.
However, our method uniquely formulates the learning procedure that has explicit consideration of operating in a fine-grained setting where highly granular OOD inputs are expected.
We also remark that the above works all consider relatively coarse-grained settings in their experiments.

\noindent \textbf{Mixing operations for OOD detection.} 
Our work also relates to, but differs significantly from a few works that utilized mixing operations in the context of OOD detection.
The work of \cite{thulasidasan_mixupood} and \cite{chun2020empirical} both directly evaluated Mixup's effect on OOD detection.
In this work, instead of plainly applying Mixup or CutMix as a regular ID training strategy, we leverage them to construct the virtual outlier distribution to expand the coverage over OOD region.
Ravikumar \etal \cite{ravikumar2020exploring} proposed to apply Mixup either between ID samples or between the training outliers.
In contrast, in our framework the mixing operations are performed between ID and outlier data, which has explicit implication in characterizing the transition from ID to OOD region.
In Sec. \ref{sec:ablation_study}, we will demonstrate the superiority of the proposed MixOE over these methods through careful ablation study.




\noindent \textbf{OOD detection in fine-grained settings.} 
As aforementioned, there are a few works that lightly considered OOD detection in fine-grained environments, yet they all have significant limitations.
In the work of \cite{perera2019deep}, a \textit{labeled} outlier dataset was used for training, which we believe is a prohibitive assumption in reality.
Even more concerning, in its experiments the training outlier dataset (ImageNet \cite{deng2009imagenet}) overlapped with the testing OOD data (CUB \cite{WahCUB_200_2011} and Stanford Dogs \cite{khosla2011novel}), which makes the detection (arguably) trivial.
The work of \cite{ahmed2020detecting} and \cite{techapanurak2021practical} are limited in that the scale of the fine-grained detection problems in their experiments is rather small, \ie, only tens of or even fewer ID classes were considered, whereas we operate at a much larger scale with hundreds of ID classes.
Later, we will also show that the method studied in \cite{ahmed2020detecting} is not yet effective for detecting fine-grained OOD examples.



We also notice some recent works that purposefully consider cases where the test OOD samples are semantically related to the ID classes \cite{winkens2020contrastive,kaur2021all}.
The critical difference here is that they did \textit{not} consider the ID classes to be granular.
We argue that without such ID granularity, it may be debatable whether a model \textit{should} be asked to detect granular OOD samples!
For example, consider an experiment in \cite{winkens2020contrastive} that regards ``leopard'' (in CIFAR-100) as a granular OOD sample for a CIFAR-10 model (which contains a ``cat'' class).
This situation begs the question: given a model trained on cats, is it more desirable to generalize to the notion of a leopard\footnote{This is actually the desired behaviour in the problem of subpopulation shift robustness \cite{santurkar2021breeds}.}, or to identify leopards as OOD?
In our work, since we consider ID classes that are themselves highly granular (\eg, unique bird species), such ambiguity is avoided and the fine-grained novelties (\eg, novel bird species) should be considered OOD.

\section{Challenges of OOD detection in fine-grained settings}
\label{sec:S3}

This section describes a detailed study on OOD detection in fine-grained settings and serves to further motivate our goal of improving detection explicitly in fine-grained scenarios.
In Sec. \ref{sec:3.1} we describe the construction of the four fine-grained test environments, which is necessary because those presented in prior works are limited in scale.
In Sec. \ref{sec:3.2}, through initial evaluations we show that fine-grained OOD detection presents distinct challenges for existing methods.

\begin{table}
  \centering
  \caption{Comparison of constructed fine-grained OOD settings. For example, in our third test environment there are 150 ID and 46 (fine-grained) OOD categories. We consider a larger scale than previous works.}
  \resizebox{.35\textwidth}{!}{
  \begin{tabular}{lll}
    \toprule
     & \# ID classes & \# OOD classes \\
    \midrule
    \cite{perera2019deep} & [100, 60] & [100, 60]\\
    \cite{ahmed2020detecting} & [11, 9, 8, 7, 7] & [1, 1, 1, 1, 1] \\
    \cite{techapanurak2021practical} & [46, 20] & [47, 5]\\
    Ours & [200, 150, 150, 90] & [55, 50, 46, 12]\\ \bottomrule
  \end{tabular}
  }
  \label{tab:test_env}
\end{table}

\subsection{Test environments}
\label{sec:3.1}

The test environments are curated from four public fine-grained visual classification (FGVC) datasets, namely FGVC-Aircraft \cite{maji13fine-grained}, Stanford Cars \cite{KrauseStarkDengFei-Fei_3DRR2013}, Butterfly \cite{chen2018butterfly}, and North American Birds \cite{van2015NAbirds}.
We refer to them as {\aircraft}, {\car}, {\butterfly}, and {\bird}, respectively.
For each dataset, we create ID/OOD splits using a holdout class method, \ie, we keep some of the categories as ID and the rest are held out from the training set and considered OOD at test time.
Note, to avoid implicit bias that might exist in each single split \cite{ahmed2020detecting}, we randomly produce three ID/OOD splits for each dataset with equal counts of ID/OOD categories.
See Appendix \ref{ap:split_details} for more details including exact splits and number of train/test images.

In Tab. \ref{tab:test_env} we present a comparison between our constructed environments and previous ones.
By considering 100+ ID classes we are operating at a larger scale which better represents a wide variety of complex real-world tasks and avoids putting any restrictive assumptions about the complexity of the ID classification task that would make our findings less scalable.
We also leave a reasonable number of classes as OOD to reflect the diversity of the open world.

Besides fine-grained novel inputs, a reliable detector should also be able to identify coarse-grained OOD data.
Here, for each dataset, we take the images from the other datasets as coarse-grained OOD samples (\eg, when \bird{} is ID, \butterfly{}, \car{}, and \aircraft{} will be considered OOD).
The detection performance is then evaluated against both fine- and coarse-grained novelties.

\subsection{Evaluating existing methods}
\label{sec:3.2}

\noindent \textbf{Setup.}
We now evaluate six state-of-the-art detectors in the constructed fine-grained environments, including three post-training scorers\footnote{We also evaluated Mahalanobis detector \cite{lee2018simple} but simply got NaN error, which aligns with \cite{hendrycks2019scaling}'s finding  that it has scalability issue.} (MSP \cite{hendrycks2017baseline}, ODIN \cite{liang2018enhancing}, and Energy \cite{liu2020energy}) and three that incorporate training-time regularizations using auxiliary outlier data (OE \cite{hendrycks2018deep}, OE with hard example mining (OE-M) \cite{chen2021atom}, and EnergyOE \cite{liu2020energy}).
In addition, we consider the method Rotation \cite{ahmed2020detecting} which was studied in a related work that mentioned fine-grained detection.
We leave implementation details of the detection methods to Sec. \ref{sec:setup} and Appendix \ref{ap:hyperparam}.

To measure detector performance, we use two common metrics \cite{lee2018simple,lee2018training,hsu2020generalized}: true negative rate at a 95\% true positive rate (TNR95) and area under the receiver operating characteristic curve (AUROC).
While AUROC is a holistic measurement obtained by varying the threshold, TNR95 indicates what portion of OOD samples could be detected when the recall of ID data is 95\%.
TNR95 is known to better separate different detectors since achieving a high TNR95 is much harder than achieving a high AUROC \cite{hsu2020generalized}.

\begin{figure}
  \centering
  \includegraphics[width=.48\textwidth]{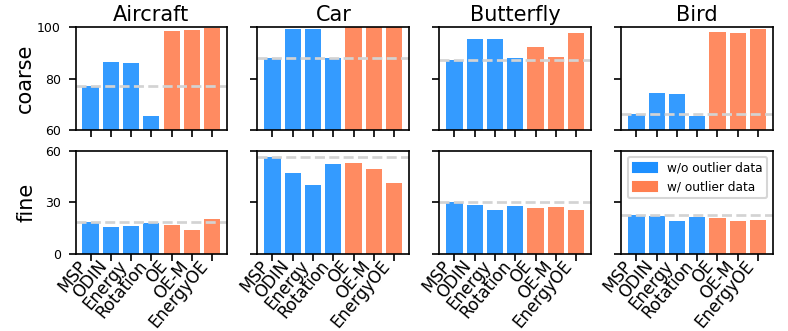}
  \caption{TNR95 of existing methods against coarse-grained (first row) and fine-grained OOD data (second row). The gray dashed line is the baseline performance (MSP). Fine-grained novelties are significantly harder to detect in all datasets for all methods. Also note how the methods that utilize outlier data ({\color{coral} coral ones}) help with coarse-grained OOD but barely improve fine-grained detection rates.}
  \label{fig:initial_eval}
  \vspace{-3mm}
\end{figure}

\noindent \textbf{Observations.}
Fig. \ref{fig:initial_eval} shows the TNR95 results on one of the three splits in each dataset (other splits present similar patterns; see Appendix \ref{ap:additional_results} Fig. \ref{fig:initial_eval_all}).
Also see the full result table including AUROC statistics in Appendix \ref{ap:additional_results} Tab. \ref{tab:full_results}.
From these results we make two important observations.
First, fine-grained OOD samples are significantly more difficult to detect than coarse-grained ones.
Specifically, while most methods can achieve more than around 80\% TNR95 when detecting coarse-grained OOD samples (Fig. \ref{fig:initial_eval} first row), the TNR95 drops to below 30\% for all methods on 3 of the 4 datasets when facing fine-grained OOD (Fig. \ref{fig:initial_eval} second row).
This observation is in line with recent findings that detection becomes more challenging when OOD data are semantically similar to ID classes \cite{hsu2020generalized,winkens2020contrastive,kaur2021all}.

Our second observation is more surprising: on the holdout fine-grained data, even the methods that explicitly include outlier data during the training (OE, OE-M, and EnergyOE) do not reliably outperform MSP, which trains the model using ID data only.
This finding directly contrasts the results on coarse-grained data, where OE/OE-M/EnergyOE consistently lead to improvements (compared with MSP).
In all, these trends clearly demonstrate that OOD detection in fine-grained settings with highly granular OOD inputs is particularly challenging for existing detectors.

\begin{figure*}[t]
  \centering
  \includegraphics[width=.95\linewidth]{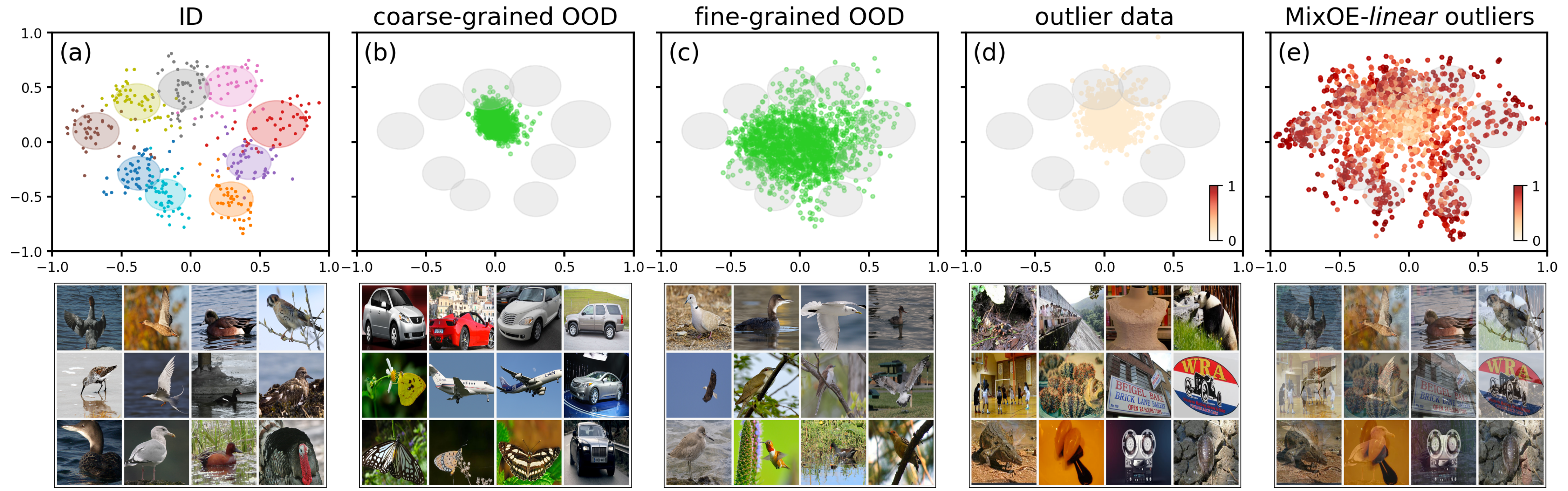}
  \caption{Visualization of the data samples (second row) and their representations in the DNN's feature space (first row). The color lightness in (d)/(e) indicates the prediction confidence encoded in the soft target of each corresponding outlier sample.
  Note, (b) and (c) are test OOD samples and are never seen during the training. 
  The empirical training outliers (d) enclose the region where coarse OOD data (b) locate but does not cover the much broader area the fine OOD samples (c) span.
  MixOE mixes the ID (a) and training outliers (d) to induce larger coverage which accounts for both coarse- and fine-grained novelties.
  Moreover, the soft targets of the mixed data will calibrate the model's prediction confidence to smoothly decay from ID to OOD.
  }
  \label{fig:method_vis}
  \vspace{-4mm}
\end{figure*}

\section{Methodology}
\label{sec:S4}

\subsection{Motivating analysis}
\label{sec:4.1}

An immediate question that arises from our initial results is \textit{why using auxiliary outlier data to explicitly regularize the model helps with detecting coarse-grained OOD samples but not fine-grained ones?}
In fact, the training outlier data we use do \textit{not} characterize/model the test coarse- or fine-grained OOD at all (see Fig. \ref{fig:method_vis}; we manually ensure this via filtering as discussed in Sec. \ref{sec:setup}).
Next, we conduct analysis to provide an explanation to this question, which also directly motivates our methodology.

The idea of our analysis is to reveal the relationship between the training outliers and test OOD samples by projecting them into the DNN's feature space.
Specifically, we forward pass the train/test OOD data into a ``standard'' pre-trained model, take the outputs of the penultimate layer as their features, and visualize them in a 2D space using the technique in \cite{pang2019rethinking} (see Appendix \ref{ap:visualize_detail} for details).
Critically, the visualization process is uniquely formed by ID data and thus remains \textit{agnostic} to OOD data.
The standard model here is trained using cross-entropy loss on ID data only.
Since OE-based methods typically fine-tune the standard model using the outlier data \cite{hendrycks2018deep,liu2020energy,chen2021atom}, by visualizing in the standard model's feature space we can anticipate/explain how the training outliers may help regularize the model.

Fig. \ref{fig:method_vis} shows the results on one of the test benches (see Appendix \ref{ap:additional_results} Fig. \ref{fig:vis_all_aircraft}-\ref{fig:vis_all_bird} for more), from which we make two key observations.
First, as shown in Fig. \ref{fig:method_vis} (b), coarse-grained OOD data locate in a rather compact region, with a small portion of samples intersecting with one of the ID clusters and others being relatively far away from the ID region.
According to Fig. \ref{fig:method_vis} (c), however, fine-grained OOD samples span a much broader area, with many of them being very close to or even within the ID clusters, due to their semantic similarity to the ID images.
This observation reinforces the TNR95 detection results in Fig. \ref{fig:initial_eval}, where fine-grained novelties are significantly more difficult to detect than coarse-grained ones.

Second, comparing Fig. \ref{fig:method_vis} (d) and (b), the auxiliary outlier data ``enclose'' the coarse OOD data region.
As a result, although  the outliers do not have any concepts related to the test coarse OOD samples (\ie, no cars/butterflies/aircrafts in Fig. \ref{fig:method_vis} (d) second row), their regularization still generalize to the test coarse OOD inputs.
On the other hand, clearly the training outliers fail to cover the larger area where many fine-grained OOD data locate, which explains why methods utilizing the outlier data have limited effect in detecting fine-grained novelties.


\subsection{Mixture Outlier Exposure}
\label{sec:4.2}

To explicitly regularize the model's behaviour in a broader region to improve detection against both coarse- and fine-grained OOD data, we would like to create a ``virtual'' outlier distribution $\Dio$ which can provide a more comprehensive coverage over the OOD region than the empirical outlier distribution $\mathcal{D}_\text{out}$.

\noindent \textbf{Generating mixed outliers.} Specifically, given an ID sample $(x_\text{in},y_\text{in})\sim\Din$ and an outlier sample $x_\text{out}\sim\mathcal{D}_\text{out}$, we propose to perform \textit{mixing operations} to generate the virtual outliers:
\begin{equation}
    \tilde{x}=\text{mix}(x_\text{in},x_\text{out},\lambda),
    \label{eq:mix_input}
\end{equation}
where $\lambda\in[0,1]$ is a coefficient controlling the contribution of each sample to the mixed one (\ie, $\lambda$ and $1-\lambda$ are the weight for $x_\text{in}$ and $x_\text{out}$, respectively).
The intuition here is directly based on our observations in Fig. \ref{fig:method_vis}: If we ``interpolate'' the ID samples (Fig. \ref{fig:method_vis} (a)) and outlier data (Fig. \ref{fig:method_vis} (d)), the resulting samples are likely to span a larger area and cover the region where fine-grained OOD data locate (Fig. \ref{fig:method_vis} (c)).

We find that simple pixel-level operations, \eg, linear interpolation \cite{zhang2018mixup} and cut-paste operation \cite{yun2019cutmix}, can already result in samples that cover the fine-grained OOD region.
Thus in this work we instantiate Eqn. \ref{eq:mix_input} with these operations and denote them as \textit{linear} mixing and \textit{cut} mixing, respectively.
However, we remark that this operation may be as simple or as complex as desired; it is not constrained to \cite{zhang2018mixup,yun2019cutmix} and allows possible extension in the future.

To demonstrate the effect of mixed samples $\tilde{x}$, we again visualize them in the DNN model's feature space in Fig. \ref{fig:method_vis} (e) (here we use \textit{linear} mixing as an example; see Appendix \ref{ap:additional_results} Fig. \ref{fig:vis_all_aircraft}-\ref{fig:vis_all_bird} for more). 
Importantly, unlike the empirical outlier distribution $\mathcal{D}_\text{out}$ in Fig. \ref{fig:method_vis} (d), our virtual outlier samples can span a larger area (being \textit{both} near to and far away from the ID clusters) by varying the coefficient $\lambda$.
As a result, we anticipate that when training with the samples from $\Dio$, the model's behaviour on both fine- and coarse-grained OOD data can be regularized.

\noindent \textbf{Assigning training targets.}
Now that we have generated mixed outliers for training, the next important step is to decide the corresponding training targets.
\textit{Our key insight here is to regularize the model such that its prediction confidence can smoothly decay as the input transitions from ID to OOD.}
To this end, we assign soft target $\tilde{y}$ corresponding to the mixed sample $\tilde{x}$ as follows:
\begin{equation}
    \tilde{y}=\lambda y_\text{in} + (1-\lambda)\mathcal{U},
    \label{eq:mix_output}
\end{equation}
where $y_\text{in}$ is the one-hot label of the ID sample $x_\text{in}$, and $\mathcal{U}$ represents the uniform distribution over the ID categories.
The prediction confidence (\ie, maximum softmax probability \cite{hendrycks2017baseline}) encoded in $\tilde{y}$ is $\tilde{y}_\text{pred. conf.}=\lambda+(1-\lambda)\frac{1}{K}$, where $K$ is the number of ID categories.
Concretely, when the mixed sample is OOD (\ie, $\lambda=0$), we force the model to be least confident/maximally uncertain on that sample (\ie, $\tilde{y}_\text{pred. conf.}=\frac{1}{K}$); when the mixed sample is ID (\ie, $\lambda=1$), the model is trained to make a confident prediction (\ie, $\tilde{y}_\text{pred. conf.}=1$).
The confidence of the intermediate mixed samples (\ie, $\lambda\in(0,1)$) is smoothly modulated by $\lambda$.
A visualization of such effect can be seen in Fig. \ref{fig:method_vis} (e), where darker points correspond to a higher confidence that is encoded in their soft targets $\tilde{y}$.
Unlike previous methods, we remark that this formulation uniquely enables the model to have better calibration over a wider range of confidence levels.
As a result, our method can have effects across a spectrum of OOD granularities, which is crucial for detectors in fine-grained environments as the OOD samples could be highly granular.

\noindent \textbf{Training objective.}
The above two technical insights make up our training algorithm, Mixture Outlier Exposure (MixOE), whose objective is formulated as
\begin{equation}
\label{eq:mixoe}
\mathbb{E}_{(x,y)\sim\Din} \big[\mathcal{L}(f(x),y)\big]+
\beta\mathbb{E}_{(\tilde{x},\tilde{y})\sim\Dio}\big[\mathcal{L}(f(\tilde{x}),\tilde{y})\big].
\end{equation}
Here, $\mathcal{L}(f(x),y)$ is the cross-entropy loss between the DNN's predicted distribution $f(x)$ and the ground truth distribution $y$, and $\beta$ is a weighting term.
During training, at each iteration the $\lambda$ in Eqn. \ref{eq:mix_input} and Eqn. \ref{eq:mix_output} is sampled from a Beta distribution $\text{Beta}(\alpha, \alpha)$ for an introduced hyperparameter $\alpha$.
Both $\alpha$ and $\beta$ can be determined using validation data (details are in Sec. \ref{sec:setup}).
We also remark here that the outlier dataset used to construct $\Dio$ is \textit{unlabeled} and does \textit{not} need to contain any data related to the test OOD data distribution.
After training, the detection will be performed by thresholding the prediction confidence since MixOE explicitly calibrates the confidence on outliers during the training.

\noindent \textbf{Relationship with prior methods.}
The training objective of the vanilla OE \cite{hendrycks2018deep} is $\mathbb{E}_{(x,y)\sim\Din} \big[\mathcal{L}(f(x),y)\big]+
\beta\mathbb{E}_{x_\text{out}\sim\mathcal{D}_\text{out}}\big[\mathcal{L}(f(x_\text{out}),\mathcal{U})\big]$, which encourages the model's output to resemble uniform distribution $\mathcal{U}$ on the empirical outliers from $\mathcal{D}_\text{out}$.
What separates MixOE from OE is the novel idea of using generated virtual outliers to expand the regularization over a broader region and controlling how the confidence decay from ID to OOD.
Also, note that MixOE can degenerate to vanilla OE if $\lambda$ is fixed to $0$ during the training.
The other two methods, OE-M \cite{chen2021atom} and EnergyOE \cite{liu2020energy}, share the same idea as OE except that \cite{chen2021atom} uses the ``hardest'' outliers from $\mathcal{D}_\text{out}$ and \cite{liu2020energy} adopts a different scoring function for detection.
They still uses only the empirical outliers and has no control over the confidence decay.
Thus we believe they share similar shortcomings to OE, which we will confirm with experimental results shortly.
\section{Experiments}
\label{sec:S5}

\subsection{Setup}
\label{sec:setup}

\noindent \textbf{Baselines.}
Same as in Sec. \ref{sec:3.2}, we consider a total of seven baseline methods, including six state-of-the-art methods \cite{hendrycks2017baseline,liang2018enhancing,liu2020energy,hendrycks2018deep,chen2021atom} and one that was studied in a previous fine-grained setting \cite{ahmed2020detecting}.
We also re-emphasize that there are very few methods that have considerations for operating in fine-grained environments as this is an underexplored topic. 



\noindent \textbf{Training details.}
For the post-training scoring methods, we train ResNet-50 models \cite{he2016deep} in the standard fashion, \ie, by minimizing the cross-entropy loss over the ID-only training dataset.
Specifically, we train the model for 90 epochs using SGD with the batch size being 32. 
Following common practices in fine-grained classification research \cite{zheng2017learning,hsu2019fine,yang2021re}, the model is initialized with ImageNet pre-trained weights.
The initial learning rate is 0.001 and is decayed by cosine learning rate schedule \cite{loshchilov2016sgdr}.
For Rotation \cite{ahmed2020detecting}, we train the model using its objective with the same setup as the standard training.

For methods that utilize auxiliary outlier data (OE, OE-M, EnergyOE, and MixOE), we only fine-tune the trained standard model with the corresponding objective for 10 epochs, following \cite{hendrycks2018deep,liu2020energy}.
Therefore, MixOE and other OE-based methods only induce marginal computation overhead.
The fine-tuning also adopts cosine schedule with the initial learning rate being 0.001.
The batch size of ID data is still 32.
For OE/OE-M/EnergyOE, as suggested in their papers, we set the batch size of outlier data to be twice as the ID batch size, which is 64.
In the case of MixOE, we keep outlier batch size same as the ID batch size.
As a result, MixOE actually uses only half of the outliers used by other methods.

\noindent \textbf{The auxiliary outlier set.}
The training outlier set $\mathcal{D}_\text{out}$ we consider is WebVision 1.0 \cite{li2017webvision}, which contains natural images crawled from Flickr and Google by querying with the 1,000 categories of ImageNet.  
We believe this dataset represents a realistic and practical construction of $\mathcal{D}_\text{out}$ for many ID tasks in the natural imagery domain.

Importantly, to avoid arguments of ``cheating'', \textit{we filter out images that are relevant to the considered OOD tasks from the outlier set based on WordNet ID} \cite{miller1995wordnet}.
Specifically, a total of 491K images related to aircraft/car/butterfly/bird are removed.
Thus, the training outliers reveal no information about the test coarse- or fine-grained OOD data (see Fig. \ref{fig:method_vis} for visualization).
After the filtering, there are 1948K images left in the outlier set.
However, since MixOE just fine-tunes the model for 10 epochs, at most 70K images are actually used during the training.


\noindent \textbf{Hyperparameter tuning.}
We take great care to ensure that the hyperparameter tuning is fair.
Concretely, we randomly holdout a portion of samples from the ID and outlier training set to serve as ID/OOD validation data.
Critically, note that the OOD validation data reveals no information about the test-time OOD distribution since we already filter out all relevant images from $\mathcal{D}_\text{out}$.


With the selected ID and OOD validation samples, we tune the hyperparameters such that the OOD detection performance is maximized and the ID classification accuracy is minimally affected.
To test the method robustness, in each of the four environments we only tune the hyperparameter once for each approach using a single split; the determined hyperparameter is then applied to all splits from the same dataset.
In Appendix \ref{ap:hyperparam} we present a detailed list of the candidate hyperparameter values and the final deteremined values we use for each method in our experiments.

\noindent \textbf{Evaluation.}
The evaluation procedure follows the one described in Sec. \ref{sec:3.1} and Sec. \ref{sec:3.2}.
For each dataset, we consider the holdout classes as fine-grained OOD data and the samples from other datasets as coarse-grained OOD data.
Following \cite{hsu2020generalized,lee2018simple,lee2018training}, we consider ID as positive and OOD as negative, and use TNR95 and AUROC as the metrics.

\subsection{Results}
\label{sec:results}

\noindent \textbf{Detection performance.}
Tab. \ref{tab:tnr_results} shows the TNR95 results across the four test benches.
The AUROC statistics yield similar patterns to TNR95 and are left in the expanded Tab. \ref{tab:full_results} in Appendix \ref{ap:additional_results}.

\begin{table}
\centering
\caption{Detection performance in terms of TNR95 statistics. The number before and after the slash is for coarse-grained and fine-grained OOD samples, respectively. Avg. diff. is the average difference (across three splits) relative to MSP. Clearly, MixOE consistently leads to notable \textcolor{\upcolor}{improvements} against both coarse- and fine-grained novelties, while other methods \textcolor{\downcolor}{degrade} the fine-grained OOD detection performance and thus do not qualify for a reliable detector in fine-grained environments.}
\resizebox{0.48\textwidth}{!}{
    \begin{tabular}{llrrrr}
    \toprule
    $\mathcal{D}_{\text{in}}$ & Method & Split 1 &Split 2 &Split 3 &Avg. diff. \\ \midrule

    \multirow{9}{*}{\rotatebox[origin=c]{90}{Aircraft}}
    &MSP \cite{hendrycks2017baseline} 
    &75.0 / 29.9  &61.6 / 15.9  &77.1 / 18.5 &- / -
    \\
    
    &ODIN \cite{liang2018enhancing}
    &87.5 / 30.2  &73.2 / 15.3  &86.5 / 15.8 &\textcolor{\upcolor}{+11.2} / \textcolor{\downcolor}{--1.0}
    \\
    
    &Energy \cite{liu2020energy} 
    &88.5 / 30.1  &74.4 / 14.6  &86.2 / 16.3 &\textcolor{\upcolor}{+11.8} / \textcolor{\downcolor}{--1.1}
    \\
    
    &Rotation \cite{ahmed2020detecting}
    &65.5 / 31.4  &55.0 / 15.9  &65.5 / 17.6 &\textcolor{\downcolor}{--9.2} / \textcolor{\upcolor}{+0.2}
    \\[1mm] \cdashline{2-6} \\[-3mm]
    
    
    &OE \cite{hendrycks2018deep}
    &99.3 / 27.8  &98.5 / 16.0  &98.7 / 16.5 &\textcolor{\upcolor}{+27.6} / \textcolor{\downcolor}{--1.3}
    \\
    
    &OE-M \cite{chen2021atom}
    &99.6 / 25.0  &98.5 / 16.0  &98.9 / 14.0 &\textcolor{\upcolor}{+27.8} / \textcolor{\downcolor}{--3.1}
    \\
    
    &EnergyOE \cite{liu2020energy} 
    &99.8 / 30.3  &99.7 / 17.0  &99.7 / 19.9 &\textcolor{\upcolor}{+28.5} / \textcolor{\upcolor}{+1.0}
    \\ 
    
    
    \rowcolor{gray!15}\cellcolor{white} &\method-\textit{linear}
    &93.2 / 41.4  &88.4 / 24.6  &92.1 / 16.5 &\textcolor{\upcolor}{+20.0} / \textcolor{\upcolor}{+6.1}
    \\
    
    
    \rowcolor{gray!15}\cellcolor{white} &\method-\textit{cut}
    &99.0 / 39.8  &99.4 / 23.7  &99.4 / 24.9 &\textcolor{\upcolor}{+28.0} / \textcolor{\upcolor}{+8.0}
    \\
    \midrule

    \multirow{9}{*}{\rotatebox[origin=c]{90}{Car}}
    &MSP \cite{hendrycks2017baseline} 
    &95.5 / 58.5  &88.0 / 56.3  &78.8 / 53.5 &- / -
    \\
    
    &ODIN \cite{liang2018enhancing}
    &99.6 / 55.6  &99.1 / 47.0  &97.8 / 49.0 &\textcolor{\upcolor}{+11.4} / \textcolor{\downcolor}{--5.6}
    \\
    
    &Energy \cite{liu2020energy} 
    &99.7 / 49.1  &99.4 / 39.7  &99.1 / 42.6 &\textcolor{\upcolor}{+12.0} / \textcolor{\downcolor}{--12.3}
    \\
    
    &Rotation \cite{ahmed2020detecting}
    &97.7 / 58.9  &88.1 / 52.4  &81.3 / 50.4 &\textcolor{\upcolor}{+1.6} / \textcolor{\downcolor}{--2.2}
    \\[1mm] \cdashline{2-6} \\[-3mm]
    
    
    &OE \cite{hendrycks2018deep}
    &99.9 / 53.2  &100.0 / 53.0  &99.9 / 51.2 &\textcolor{\upcolor}{+12.5} / \textcolor{\downcolor}{--3.6}
    \\
    
    &OE-M \cite{chen2021atom}
    &99.9 / 53.6  &100.0 / 49.4  &100.0 / 50.6 &\textcolor{\upcolor}{+12.5} / \textcolor{\downcolor}{--4.9}
    \\
    
    &EnergyOE \cite{liu2020energy} 
    &100.0 / 52.6  &100.0 / 41.0  &100.0 / 44.9 &\textcolor{\upcolor}{+12.6} / \textcolor{\downcolor}{--9.9}
    \\ 
    
    
    \rowcolor{gray!15}\cellcolor{white} &\method-\textit{linear}
    &99.6 / 65.9  &99.7 / 62.9  &99.5 / 60.1 &\textcolor{\upcolor}{+12.2} / \textcolor{\upcolor}{+6.9}
    \\
    
    
    \rowcolor{gray!15}\cellcolor{white} &\method-\textit{cut}
    &99.9 / 70.3  &100.0 / 69.8  &99.9 / 66.5 &\textcolor{\upcolor}{+12.5} / \textcolor{\upcolor}{+12.8}
    \\
    \midrule

    \multirow{9}{*}{\rotatebox[origin=c]{90}{Butterfly}}
    &MSP \cite{hendrycks2017baseline} 
    &87.1 / 29.9  &89.9 / 31.8  &88.4 / 36.6 &- / -
    \\
    
    &ODIN \cite{liang2018enhancing}
    &95.2 / 28.2  &95.5 / 32.5  &95.6 / 38.7 &\textcolor{\upcolor}{+7.0} / \textcolor{\upcolor}{+0.4}
    \\
    
    &Energy \cite{liu2020energy} 
    &95.3 / 25.5  &95.2 / 30.2  &95.6 / 36.1 &\textcolor{\upcolor}{+6.9} / \textcolor{\downcolor}{--2.2}
    \\
    
    &Rotation \cite{ahmed2020detecting}
    &87.9 / 27.6  &88.5 / 31.2  &86.2 / 37.0 &\textcolor{\downcolor}{--0.9} / \textcolor{\downcolor}{--0.8}
    \\[1mm] \cdashline{2-6} \\[-3mm]
    
    
    &OE \cite{hendrycks2018deep}
    &92.2 / 26.5  &93.7 / 32.1  &94.3 / 34.3 &\textcolor{\upcolor}{+4.9} / \textcolor{\downcolor}{--1.8}
    \\
    
    &OE-M \cite{chen2021atom}
    &99.9 / 53.6  &100.0 / 49.4  &100.0 / 50.6 &\textcolor{\upcolor}{+12.5} / \textcolor{\downcolor}{--4.9}
    \\
    
    &EnergyOE \cite{liu2020energy} 
    &97.8 / 25.1  &96.9 / 30.5  &98.2 / 37.2 &\textcolor{\upcolor}{+9.2} / \textcolor{\downcolor}{--1.8}
    \\ 
    
    
    \rowcolor{gray!15}\cellcolor{white} &\method-\textit{linear}
    &95.3 / 32.6  &93.9 / 37.9  &95.5 / 45.0 &\textcolor{\upcolor}{+6.4} / \textcolor{\upcolor}{+5.7}
    \\
    
    
    \rowcolor{gray!15}\cellcolor{white} &\method-\textit{cut}
    &94.9 / 35.8  &94.1 / 38.8  &92.7 / 46.0 &\textcolor{\upcolor}{+5.4} / \textcolor{\upcolor}{+7.4}
    \\
    \midrule

    \multirow{9}{*}{\rotatebox[origin=c]{90}{Bird}}
    &MSP \cite{hendrycks2017baseline} 
    &72.3 / 22.6  &67.4 / 22.3  &66.4 / 22.3 &- / -
    \\
    
    &ODIN \cite{liang2018enhancing}
    &80.9 / 22.7  &77.2 / 21.5  &74.3 / 21.9 &\textcolor{\upcolor}{+8.8} / \textcolor{\downcolor}{--0.4}
    \\
    
    &Energy \cite{liu2020energy} 
    &80.8 / 20.3  &76.5 / 18.4  &73.9 / 18.8 &\textcolor{\upcolor}{+8.4} / \textcolor{\downcolor}{--3.2}
    \\
    
    &Rotation \cite{ahmed2020detecting}
    &71.3 / 23.6  &64.0 / 24.0  &65.4 / 21.5 &\textcolor{\downcolor}{--1.8} / \textcolor{\upcolor}{+0.6}
    \\[1mm] \cdashline{2-6} \\[-3mm]
    
    
    &OE \cite{hendrycks2018deep}
    &98.2 / 20.6  &97.9 / 22.9  &97.9 / 20.7 &\textcolor{\upcolor}{+29.3} / \textcolor{\downcolor}{--1.0}
    \\
    
    &OE-M \cite{chen2021atom}
    &98.7 / 19.8  &98.7 / 21.4  &97.7 / 19.2 &\textcolor{\upcolor}{+29.7} / \textcolor{\downcolor}{--2.3}
    \\
    
    &EnergyOE \cite{liu2020energy} 
    &98.6 / 19.4  &99.0 / 18.4  &99.3 / 19.5 &\textcolor{\upcolor}{+30.3} / \textcolor{\downcolor}{--3.3}
    \\ 
    
    \rowcolor{gray!15}\cellcolor{white} &\method-\textit{linear}
    &88.6 / 24.9  &83.9 / 26.7  &86.3 / 28.6 &\textcolor{\upcolor}{+17.6} / \textcolor{\upcolor}{+4.3}
    \\
    
    \rowcolor{gray!15}\cellcolor{white} &\method-\textit{cut}
    &91.0 / 27.7  &91.8 / 24.6  &92.9 / 27.7 &\textcolor{\upcolor}{+23.2} / \textcolor{\upcolor}{+4.3}
    \\
    \bottomrule
    \end{tabular}
    }
\label{tab:tnr_results}
\end{table}

Our first observation is that MixOE consistently achieves the best detection performance against fine-grained OOD samples.
Specifically, averaged across the three splits on the [\aircraft{}, \car{}, \butterfly{}, \bird{}] tasks, MixOE-\textit{linear} and MixOE-\textit{cut} improve the TNR95 over MSP by [+6.1\%, +6.9\%, +5.7\%, +4.3\%] and [+8.0\%, +12.8\%, +7.4\%, +4.3\%], respectively.
In comparison, vanilla OE's relative improvement over MSP is [--1.3\%, --3.6\%, --1.8\%, --1.0\%].
Similar to OE, the other baseline methods also result in performance degradation in the face of fine-grained OOD on many (if not all) datasets.
We also remark that MixOE-\textit{linear} and MixOE-\textit{cut} both lead to significant improvements, demonstrating that the validity of the idea behind MixOE is independent of the specific mixing operations being used.

Our second observation is that MixOE can perform on par with state-of-the-art methods in detecting coarse-grained OOD samples.
On the four tasks MixOE-\textit{linear} and MixOE-\textit{cut} improve the TNR95 over MSP by [+20.0\%, +12.2\%, +6.4\%, +17.6\%] and [+28.0\%, +12.5\%, +5.4\%, +23.2\%], respectively.
OE, whose effect is exclusively towards coarse-grained OOD data, leads to improvements of [+27.6\%, +12.5\%, +4.9\%, +29.3\%].

Finally, we note that MixOE is the only method that consistently improves upon MSP against both fine- and coarse-grained novelties across all the datasets.
This unique capability of MixOE to remain effective across a spectrum of OOD granularities is critical for systems that operate in fine-grained environments, as the novel inputs during inference can be either coarse or fine.
Overall, these evaluation results display the effectiveness of MixOE for building up reliable OOD detectors in real-world fine-grained settings.

\begin{figure}[!t]
  \centering
  \includegraphics[width=.98\linewidth]{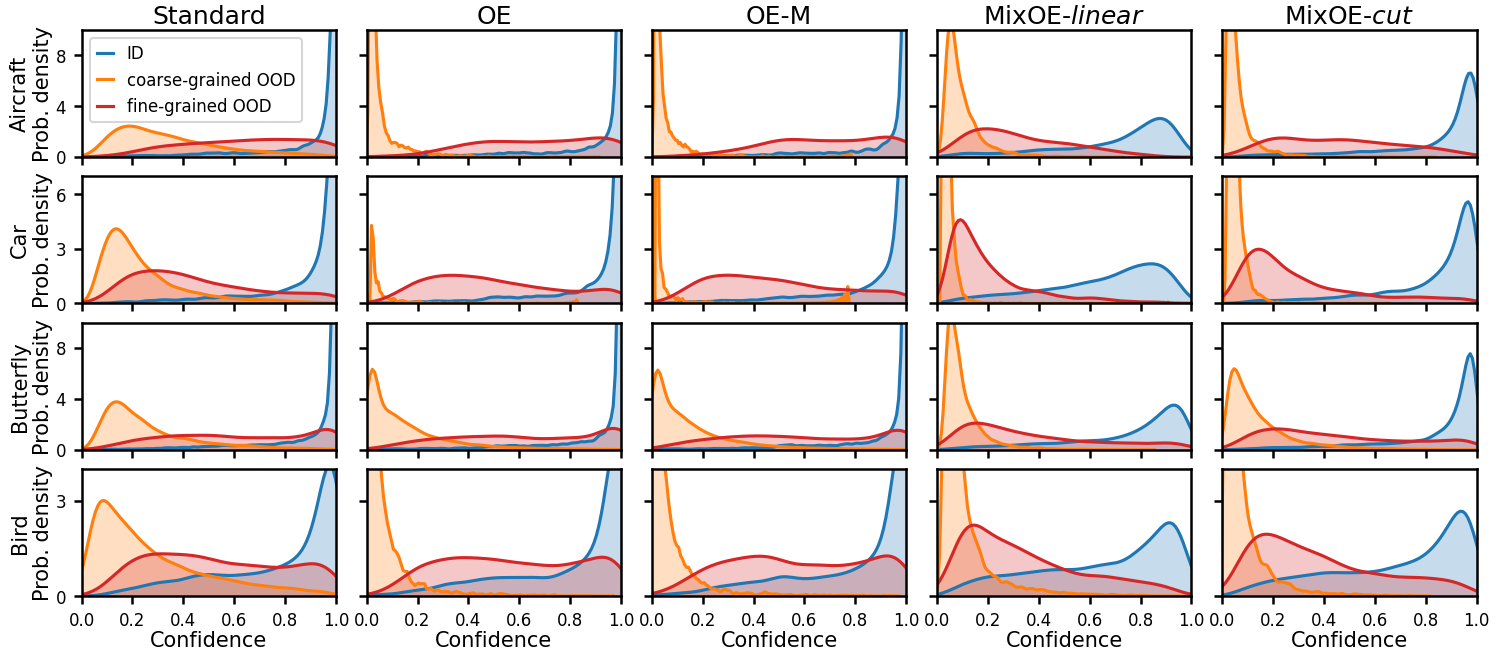}
  \caption{Comparison of the prediction confidence's distribution between methods. MixOE leads to clearer separation between the confidence of ID and OOD samples (especially fine-grained ones) and thus enables better detection.}
  \label{fig:conf_density_all}
\end{figure}

\noindent \textbf{Prediction confidences.}
To more closely understand how MixOE improves the OOD detection, we monitor the confidence distributions of the models on ID/OOD samples as MixOE functions by calibrating the predication confidence.
Fig. \ref{fig:conf_density_all} shows the probability density plots of prediction confidence on Standard, OE, OE-M, and MixOE models on one of the splits from each dataset (see Appendix \ref{ap:additional_results} Fig. \ref{fig:conf_all_aircraft}-\ref{fig:conf_all_bird} for more).
From Fig. \ref{fig:conf_density_all} we can clearly identify that, regardless of the mixing operation, the MixOE models consistently produce lower confidence predictions on fine-grained OOD samples, making them more distinguishable from ID inputs.
This observation confirms that the virtual outlier data and their corresponding soft targets introduced in MixOE indeed help regularize the model's outputs on fine-grained OOD samples.

\begin{table}
\centering
\caption{Accuracy comparison of the training algorithms. The numbers in the parenthesis are the differences relative to the accuracy of standard training. Avg. diff. shows the improvements averaged across the four datasets. Unlike other methods that \textcolor{\downcolor}{tradeoff} accuracy for detection, MixOE acutally \textcolor{\upcolor}{improves} the accuracy.}
\resizebox{0.47\textwidth}{!}{
\begin{tabular}{lccccc}
\toprule
 &Aircraft &Car &Butterfly &Bird &Avg. diff. \\ \midrule

Rotation \cite{ahmed2020detecting}
&88.5 (\textcolor{\downcolor}{--1.1}) 
&91.3 (\textcolor{\downcolor}{--0.6}) 
&88.8 (\textcolor{\downcolor}{--0.1})
&82.0 (\textcolor{\downcolor}{--0.1})
&\textcolor{\downcolor}{--0.5}
\\


OE \cite{hendrycks2018deep}
&89.2 (\textcolor{\downcolor}{--0.5})
&91.6 (\textcolor{\downcolor}{--0.2})
&88.1 (\textcolor{\downcolor}{--0.9})
&82.4 (\textcolor{\upcolor}{+0.3})
&\textcolor{\downcolor}{--0.3}
\\

OE-M \cite{chen2021atom}
&89.3 (\textcolor{\downcolor}{--0.4})
&91.1 (\textcolor{\downcolor}{--0.8})
&88.2 (\textcolor{\downcolor}{--0.8})
&82.7 (\textcolor{\upcolor}{+0.6})
&\textcolor{\downcolor}{--0.4}
\\

EnergyOE \cite{liu2020energy} 
&89.3 (\textcolor{\downcolor}{--0.3}) 
&91.8 (\textcolor{\upcolor}{--0.0}) 
&88.8 (\textcolor{\downcolor}{--0.2})
&82.3 (\textcolor{\upcolor}{+0.2})
&\textcolor{\downcolor}{--0.1}
\\ 

\rowcolor{gray!15}
\method-\textit{linear}
&90.5 (\textcolor{\upcolor}{+0.8})
&92.9 (\textcolor{\upcolor}{+1.1})
&89.3 (\textcolor{\upcolor}{+0.3})
&83.4 (\textcolor{\upcolor}{+1.3})
&\textcolor{\upcolor}{+0.9}
\\

\rowcolor{gray!15} \method-\textit{cut}
&90.1 (\textcolor{\upcolor}{+0.5})
&92.9 (\textcolor{\upcolor}{+1.1})
&90.1 (\textcolor{\upcolor}{+1.2})
&83.5 (\textcolor{\upcolor}{+1.4})
&\textcolor{\upcolor}{+1.1}
\\
\bottomrule

\end{tabular}
}
\label{tab:accuracy_results}
\end{table}

\noindent \textbf{ID classification accuracy.}
Lastly, we examine how the training methods can affect the ID classification accuracy because we do not intend to tradeoff accuracy for detection performance.
In Tab. \ref{tab:accuracy_results} we show the accuracy on each dataset averaged across the three splits (see Appendix \ref{ap:additional_results} Tab. \ref{tab:accuracy_full_results} for unaveraged results).
Interestingly, unlike other training strategies, MixOE-\textit{linear} and MixOE-\textit{cut} can improve the accuracy by 0.9\% and 1.1\% on average across the four environments, respectively.
Our hypothesis here is that since the fine-grained datasets often have relatively small number of training samples (\eg, tens of images per class), some of the training ``outliers'' generated by MixOE that are close to the ID clusters actually serve as augmented data and thus contribute to the ID accuracy.

\subsection{Ablation study}
\label{sec:ablation_study}

\noindent \textbf{MixOE v.s. Mix.}
Recall that the core idea of MixOE is to mix ID and outlier data for DNN training.
In Sec. \ref{sec:results} we have shown that this concept is more beneficial than using the outlier data alone without mixing (vanilla OE).
Here we ablate MixOE along another direction: We contrast MixOE with vanilla Mix training \cite{zhang2018mixup,yun2019cutmix} which does not use the auxiliary outlier data, \ie, mixing only ID data.
Concretely, for Mix training the generation of virtual outlier samples in Eqn. \ref{eq:mixoe} is changed to $(\tilde{x},\tilde{y})=(\text{mix}(x_1,x_2,\lambda),\lambda y_1+(1-\lambda)y_2)$, where $(x_1,y_1),(x_2,y_2)\sim\Din$.
The hyperparameter tuning procedure and training setup is the same as those for MixOE.

\begin{figure}[!t]
  \centering
  \includegraphics[width=.95\linewidth]{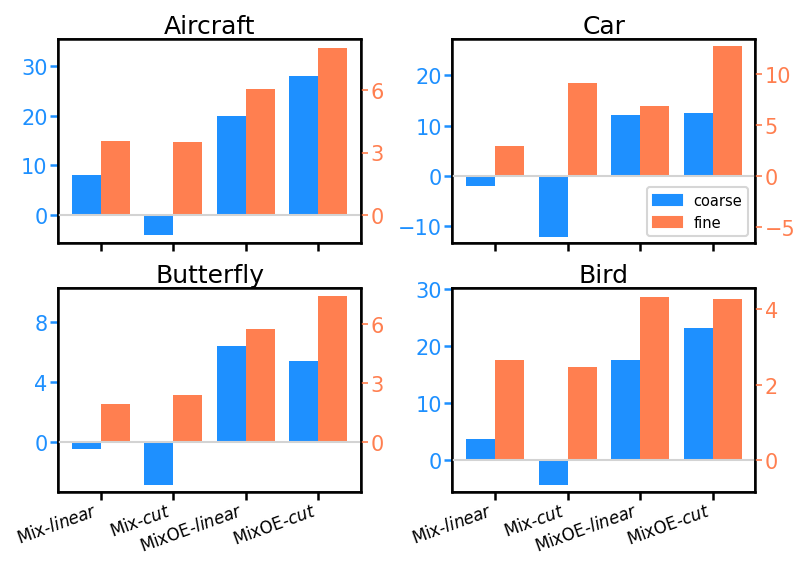}
  \caption{Comparison between MixOE and vanilla Mix training (without using outlier data) in terms of average difference in TNR95 relative to MSP. MixOE outperforms Mix against both coarse-/fine-grained OOD data.} 
  \label{fig:ablation}
  \vspace{-5mm}
\end{figure}

The comparison is presented in Fig. \ref{fig:ablation}, where we show the methods' average improvements in TNR95 relative to the baseline MSP across the three splits on each dataset.
We find that Mix training is able to provide performance gains in detecting fine-grained OOD data, but the gains are smaller than those brought by MixOE;
meanwhile, Mix training rarely improves the coarse-grained OOD detection rates.

\noindent \textbf{MixOE v.s. Mix + OE.}
Finally, we investigate whether naively combining the Mix and OE objective together can achieve similar effect to MixOE.
A detailed analysis is given in Appendix \ref{ap:ablation_study_analysis}.
The takeaway here is that Mix + OE will lead to manifold intrusion \cite{guo2019mixup}, where the training will assign distinct targets to inputs that are close to each other in the DNN's feature space, causing significant learning difficulty for the model.
Indeed, we find that when combining Mix and OE together, the model's accuracy can decrease by up to 10\%, and the TNR95 can be worse than MSP by 10\% and 20\% against coarse- and fine-grained OOD samples, respectively.
The results clearly demonstrate that \textit{MixOE's formulation is unique, effective, and cannot be replaced by a simple/naive combination of two existing objectives.}

\section{Conclusion}

In this work, we propose Mixture Outlier Exposure, a DNN training algorithm for OOD detection in fine-grained environments.
MixOE explicitly expand the coverage over the broad OOD region by mixing ID data and training outlier samples.
The mixed samples are used to regularize the model's behaviour such that the prediction confidence smoothly decays when the inputs transition from ID to OOD.
Experimental results in the four newly constructed large-scale fine-grained environments demonstrate that MixOE is able to improve detection rates against both coarse- and fine-grained OOD samples, while other methods hardly help with fine-grained detection.
We hope that this work will facilitate and inspire future research on OOD detection in the challenging fine-grained settings.

\section*{Acknowledgement}
This work is supported by FA8750-21-1-1015 and NSF-2140247.

\clearpage

{\small
\bibliographystyle{ieee_fullname}
\bibliography{egbib}

\begin{thebibliography}{10}\itemsep=-1pt

\bibitem{ahmed2020detecting}
Faruk Ahmed and Aaron Courville.
\newblock Detecting semantic anomalies.
\newblock In {\em Proceedings of the AAAI Conference on Artificial
  Intelligence}, volume~34, pages 3154--3162, 2020.

\bibitem{chen2021atom}
Jiefeng Chen, Yixuan Li, Xi Wu, Yingyu Liang, and Somesh Jha.
\newblock Atom: Robustifying out-of-distribution detection using outlier
  mining.
\newblock In {\em Joint European Conference on Machine Learning and Knowledge
  Discovery in Databases}, pages 430--445. Springer, 2021.

\bibitem{chen2018butterfly}
Tianshui Chen, Wenxi Wu, Yuefang Gao, Le Dong, Xiaonan Luo, and Liang Lin.
\newblock Fine-grained representation learning and recognition by exploiting
  hierarchical semantic embedding.
\newblock In {\em Proceedings of the 26th ACM international conference on
  Multimedia}, pages 2023--2031, 2018.

\bibitem{chun2020empirical}
Sanghyuk Chun, Seong~Joon Oh, Sangdoo Yun, Dongyoon Han, Junsuk Choe, and
  Youngjoon Yoo.
\newblock An empirical evaluation on robustness and uncertainty of
  regularization methods.
\newblock {\em arXiv preprint arXiv:2003.03879}, 2020.

\bibitem{deng2009imagenet}
Jia Deng, Wei Dong, Richard Socher, Li-Jia Li, Kai Li, and Li Fei-Fei.
\newblock Imagenet: A large-scale hierarchical image database.
\newblock In {\em 2009 IEEE conference on computer vision and pattern
  recognition}, pages 248--255. Ieee, 2009.

\bibitem{du2022towards}
Xuefeng Du, Zhaoning Wang, Mu Cai, and Sharon Li.
\newblock Towards unknown-aware learning with virtual outlier synthesis.
\newblock In {\em International Conference on Learning Representations}, 2022.

\bibitem{fan2020microscopic}
Mengran Fan, Tapabrata Chakraborti, I Eric, Chao Chang, Yan Xu, and Jens
  Rittscher.
\newblock Microscopic fine-grained instance classification through deep
  attention.
\newblock In {\em International Conference on Medical Image Computing and
  Computer-Assisted Intervention}, pages 490--499. Springer, 2020.

\bibitem{guo2019mixup}
Hongyu Guo, Yongyi Mao, and Richong Zhang.
\newblock Mixup as locally linear out-of-manifold regularization.
\newblock In {\em Proceedings of the AAAI Conference on Artificial
  Intelligence}, volume~33, pages 3714--3722, 2019.

\bibitem{he2016deep}
Kaiming He, Xiangyu Zhang, Shaoqing Ren, and Jian Sun.
\newblock Deep residual learning for image recognition.
\newblock In {\em Proceedings of the IEEE conference on computer vision and
  pattern recognition}, pages 770--778, 2016.

\bibitem{hendrycks2019scaling}
Dan Hendrycks, Steven Basart, Mantas Mazeika, Mohammadreza Mostajabi, Jacob
  Steinhardt, and Dawn Song.
\newblock Scaling out-of-distribution detection for real-world settings.
\newblock {\em arXiv preprint arXiv:1911.11132}, 2019.

\bibitem{hendrycks2017baseline}
Dan Hendrycks and Kevin Gimpel.
\newblock A baseline for detecting misclassified and out-of-distribution
  examples in neural networks.
\newblock In {\em International Conference on Learning Representations}, 2017.

\bibitem{hendrycks2018deep}
Dan Hendrycks, Mantas Mazeika, and Thomas Dietterich.
\newblock Deep anomaly detection with outlier exposure.
\newblock In {\em International Conference on Learning Representations}, 2018.

\bibitem{hsu2019fine}
Yen-Chi Hsu, Cheng-Yao Hong, Ding-Jie Chen, Ming-Sui Lee, Davi Geiger, and
  Tyng-Luh Liu.
\newblock Fine-grained visual recognition with batch confusion norm.
\newblock {\em arXiv preprint arXiv:1910.12423}, 2019.

\bibitem{hsu2020generalized}
Yen-Chang Hsu, Yilin Shen, Hongxia Jin, and Zsolt Kira.
\newblock Generalized odin: Detecting out-of-distribution image without
  learning from out-of-distribution data.
\newblock In {\em Proceedings of the IEEE/CVF Conference on Computer Vision and
  Pattern Recognition}, pages 10951--10960, 2020.

\bibitem{huang2021importance}
Rui Huang, Andrew Geng, and Yixuan Li.
\newblock On the importance of gradients for detecting distributional shifts in
  the wild.
\newblock {\em arXiv preprint arXiv:2110.00218}, 2021.

\bibitem{inkawhich2021advoe}
Nathan Inkawhich, Eric Davis, Matthew Inkawhich, Uttam Majumder, and Yiran
  Chen.
\newblock Training sar-atr models for reliable operation in open-world
  environments.
\newblock {\em IEEE Journal of Selected Topics in Applied Earth Observations
  and Remote Sensing}, 14:3954--3966, 2021.

\bibitem{kaur2021all}
Ramneet Kaur, Susmit Jha, Anirban Roy, Oleg Sokolsky, and Insup Lee.
\newblock Are all outliers alike? on understanding the diversity of outliers
  for detecting oods.
\newblock {\em arXiv preprint arXiv:2103.12628}, 2021.

\bibitem{vos_opinion}
Umar Khalid.
\newblock The method is not scalable., 2022.

\bibitem{khosla2011novel}
Aditya Khosla, Nityananda Jayadevaprakash, Bangpeng Yao, and Fei-Fei Li.
\newblock Novel dataset for fine-grained image categorization: Stanford dogs.
\newblock In {\em Proc. CVPR Workshop on Fine-Grained Visual Categorization
  (FGVC)}, volume~2. Citeseer, 2011.

\bibitem{KrauseStarkDengFei-Fei_3DRR2013}
Jonathan Krause, Michael Stark, Jia Deng, and Li Fei-Fei.
\newblock 3d object representations for fine-grained categorization.
\newblock In {\em 4th International IEEE Workshop on 3D Representation and
  Recognition (3dRR-13)}, Sydney, Australia, 2013.

\bibitem{lee2018training}
Kimin Lee, Honglak Lee, Kibok Lee, and Jinwoo Shin.
\newblock Training confidence-calibrated classifiers for detecting
  out-of-distribution samples.
\newblock In {\em International Conference on Learning Representations}, 2018.

\bibitem{lee2018simple}
Kimin Lee, Kibok Lee, Honglak Lee, and Jinwoo Shin.
\newblock A simple unified framework for detecting out-of-distribution samples
  and adversarial attacks.
\newblock In {\em NeurIPS}, 2018.

\bibitem{li2014medical}
Qing Li, Weidong Cai, Xiaogang Wang, Yun Zhou, David~Dagan Feng, and Mei Chen.
\newblock Medical image classification with convolutional neural network.
\newblock In {\em 2014 13th international conference on control automation
  robotics \& vision (ICARCV)}, pages 844--848. IEEE, 2014.

\bibitem{li2017webvision}
Wen Li, Limin Wang, Wei Li, Eirikur Agustsson, and Luc Van~Gool.
\newblock Webvision database: Visual learning and understanding from web data.
\newblock {\em arXiv preprint arXiv:1708.02862}, 2017.

\bibitem{liang2018enhancing}
Shiyu Liang, Yixuan Li, and R Srikant.
\newblock Enhancing the reliability of out-of-distribution image detection in
  neural networks.
\newblock In {\em International Conference on Learning Representations}, 2018.

\bibitem{liu2020energy}
Weitang Liu, Xiaoyun Wang, John Owens, and Yixuan Li.
\newblock Energy-based out-of-distribution detection.
\newblock {\em Advances in Neural Information Processing Systems}, 33, 2020.

\bibitem{loshchilov2016sgdr}
Ilya Loshchilov and Frank Hutter.
\newblock Sgdr: Stochastic gradient descent with warm restarts.
\newblock {\em arXiv preprint arXiv:1608.03983}, 2016.

\bibitem{maji13fine-grained}
S. Maji, J. Kannala, E. Rahtu, M. Blaschko, and A. Vedaldi.
\newblock Fine-grained visual classification of aircraft.
\newblock Technical report, 2013.

\bibitem{miller1995wordnet}
George~A Miller.
\newblock Wordnet: a lexical database for english.
\newblock {\em Communications of the ACM}, 38(11):39--41, 1995.

\bibitem{pang2019rethinking}
Tianyu Pang, Kun Xu, Yinpeng Dong, Chao Du, Ning Chen, and Jun Zhu.
\newblock Rethinking softmax cross-entropy loss for adversarial robustness.
\newblock In {\em International Conference on Learning Representations}, 2019.

\bibitem{perera2019deep}
Pramuditha Perera and Vishal~M Patel.
\newblock Deep transfer learning for multiple class novelty detection.
\newblock In {\em Proceedings of the IEEE/CVF Conference on Computer Vision and
  Pattern Recognition}, pages 11544--11552, 2019.

\bibitem{radford2015unsupervised}
Alec Radford, Luke Metz, and Soumith Chintala.
\newblock Unsupervised representation learning with deep convolutional
  generative adversarial networks.
\newblock {\em arXiv preprint arXiv:1511.06434}, 2015.

\bibitem{ravikumar2020exploring}
Deepak Ravikumar, Sangamesh Kodge, Isha Garg, and Kaushik Roy.
\newblock Exploring vicinal risk minimization for lightweight
  out-of-distribution detection.
\newblock {\em arXiv preprint arXiv:2012.08398}, 2020.

\bibitem{roady2020open}
Ryne Roady, Tyler~L Hayes, Ronald Kemker, Ayesha Gonzales, and Christopher
  Kanan.
\newblock Are open set classification methods effective on large-scale
  datasets?
\newblock {\em Plos one}, 15(9):e0238302, 2020.

\bibitem{santurkar2021breeds}
Shibani Santurkar, Dimitris Tsipras, and Aleksander Madry.
\newblock {\{}BREEDS{\}}: Benchmarks for subpopulation shift.
\newblock In {\em International Conference on Learning Representations}, 2021.

\bibitem{sastry2020detecting}
Chandramouli~Shama Sastry and Sageev Oore.
\newblock Detecting out-of-distribution examples with gram matrices.
\newblock In {\em International Conference on Machine Learning}, pages
  8491--8501. PMLR, 2020.

\bibitem{tack2020csi}
Jihoon Tack, Sangwoo Mo, Jongheon Jeong, and Jinwoo Shin.
\newblock Csi: Novelty detection via contrastive learning on distributionally
  shifted instances.
\newblock {\em Advances in Neural Information Processing Systems}, 33, 2020.

\bibitem{techapanurak2021bridging}
Engkarat Techapanurak, Anh-Chuong Dang, and Takayuki Okatani.
\newblock Bridging in-and out-of-distribution samples for their better
  discriminability.
\newblock {\em arXiv preprint arXiv:2101.02500}, 2021.

\bibitem{techapanurak2021practical}
Engkarat Techapanurak and Takayuki Okatani.
\newblock Practical evaluation of out-of-distribution detection methods for
  image classification.
\newblock {\em arXiv preprint arXiv:2101.02447}, 2021.

\bibitem{thulasidasan_mixupood}
Sunil Thulasidasan, Gopinath Chennupati, Jeff~A Bilmes, Tanmoy Bhattacharya,
  and Sarah Michalak.
\newblock On mixup training: Improved calibration and predictive uncertainty
  for deep neural networks.
\newblock In H. Wallach, H. Larochelle, A. Beygelzimer, F. d\textquotesingle
  Alch\'{e}-Buc, E. Fox, and R. Garnett, editors, {\em Advances in Neural
  Information Processing Systems}, volume~32. Curran Associates, Inc., 2019.

\bibitem{van2015NAbirds}
Grant Van~Horn, Steve Branson, Ryan Farrell, Scott Haber, Jessie Barry, Panos
  Ipeirotis, Pietro Perona, and Serge Belongie.
\newblock Building a bird recognition app and large scale dataset with citizen
  scientists: The fine print in fine-grained dataset collection.
\newblock In {\em Proceedings of the IEEE Conference on Computer Vision and
  Pattern Recognition}, pages 595--604, 2015.

\bibitem{WahCUB_200_2011}
C. Wah, S. Branson, P. Welinder, P. Perona, and S. Belongie.
\newblock {The Caltech-UCSD Birds-200-2011 Dataset}.
\newblock Technical Report CNS-TR-2011-001, California Institute of Technology,
  2011.

\bibitem{winkens2020contrastive}
Jim Winkens, Rudy Bunel, Abhijit~Guha Roy, Robert Stanforth, Vivek Natarajan,
  Joseph~R Ledsam, Patricia MacWilliams, Pushmeet Kohli, Alan Karthikesalingam,
  Simon Kohl, et~al.
\newblock Contrastive training for improved out-of-distribution detection.
\newblock {\em arXiv preprint arXiv:2007.05566}, 2020.

\bibitem{xia2018dota}
Gui-Song Xia, Xiang Bai, Jian Ding, Zhen Zhu, Serge Belongie, Jiebo Luo, Mihai
  Datcu, Marcello Pelillo, and Liangpei Zhang.
\newblock Dota: A large-scale dataset for object detection in aerial images.
\newblock In {\em Proceedings of the IEEE Conference on Computer Vision and
  Pattern Recognition}, pages 3974--3983, 2018.

\bibitem{xia2017aid}
Gui-Song Xia, Jingwen Hu, Fan Hu, Baoguang Shi, Xiang Bai, Yanfei Zhong,
  Liangpei Zhang, and Xiaoqiang Lu.
\newblock Aid: A benchmark data set for performance evaluation of aerial scene
  classification.
\newblock {\em IEEE Transactions on Geoscience and Remote Sensing},
  55(7):3965--3981, 2017.

\bibitem{yang2021re}
Shaokang Yang, Shuai Liu, Cheng Yang, and Changhu Wang.
\newblock Re-rank coarse classification with local region enhanced features for
  fine-grained image recognition.
\newblock {\em arXiv preprint arXiv:2102.09875}, 2021.

\bibitem{yun2019cutmix}
Sangdoo Yun, Dongyoon Han, Seong~Joon Oh, Sanghyuk Chun, Junsuk Choe, and
  Youngjoon Yoo.
\newblock Cutmix: Regularization strategy to train strong classifiers with
  localizable features.
\newblock In {\em Proceedings of the IEEE/CVF International Conference on
  Computer Vision}, pages 6023--6032, 2019.

\bibitem{zhang2018mixup}
Hongyi Zhang, Moustapha Cisse, Yann~N. Dauphin, and David Lopez-Paz.
\newblock mixup: Beyond empirical risk minimization.
\newblock In {\em International Conference on Learning Representations}, 2018.

\bibitem{zheng2017learning}
Heliang Zheng, Jianlong Fu, Tao Mei, and Jiebo Luo.
\newblock Learning multi-attention convolutional neural network for
  fine-grained image recognition.
\newblock In {\em Proceedings of the IEEE international conference on computer
  vision}, pages 5209--5217, 2017.

\bibitem{zhu2017deep}
Xiao~Xiang Zhu, Devis Tuia, Lichao Mou, Gui-Song Xia, Liangpei Zhang, Feng Xu,
  and Friedrich Fraundorfer.
\newblock Deep learning in remote sensing: A comprehensive review and list of
  resources.
\newblock {\em IEEE Geoscience and Remote Sensing Magazine}, 5(4):8--36, 2017.

\end{thebibliography}
}

\appendix
\onecolumn

\section{Details of the constructed ID/OOD splits}
\label{ap:split_details}

\begin{table*}[!ht]
\centering
\caption{The number of images in each constructed split.}
\begin{tabular}{p{1.7cm}p{1.4cm}cccc}
\toprule

& & \multicolumn{3}{c}{ID}
& \multirow{2}{*}{(Fine-grained) OOD}
\\ \cmidrule(lr){3-5}

& & train & validation & test
\\\midrule

\multirow{3}{*}{\aircraft{}}
& Split 1 & 5396 & 604 & 3000 & 333
\\
& Split 2 & 5380 & 618 & 3002 & 331
\\
& Split 3 & 5400 & 597 & 3003 & 330
\\
\midrule

\multirow{3}{*}{\car{}}
& Split 1 & 5664 & 623 & 6210 & 1831
\\
& Split 2 & 5657 & 625 & 6197 & 1844
\\
& Split 3 & 5574 & 633 & 6131 & 1910
\\
\midrule

\multirow{3}{*}{\butterfly{}}
& Split 1 & 7072 & 798 & 11513 & 3458
\\
& Split 2 & 6813 & 735 & 11019 & 3952
\\
& Split 3 & 6941 & 743 & 11219 & 3752
\\
\midrule

\multirow{3}{*}{\bird{}}
& Split 1 & 7377 & 809 & 8547 & 2149
\\
& Split 2 & 7278 & 812 & 8449 & 2247
\\
& Split 3 & 7219 & 798 & 8446 & 2250
\\

\bottomrule

\end{tabular}
\label{tab:split_details}
\end{table*}

In Tab. \ref{tab:split_details} we list the number of train/validation/test images in each ID split and the number of holdout images in each holdout (fine-grained) OOD split.
The exact classes in each split is presented below.
Note, the \aircraft{} and \car{} dataset do not have explicit class identifier, so we create numerical index ourselves.
Please see our code for details.

\begin{itemize}
    \item \aircraft{}
    \begin{itemize}
        \item Split 1
        \begin{itemize}
            \item ID: 26, 86, 2, 55, 75, 93, 16, 73, 54, 95, 53, 92, 78, 13, 7, 30, 22, 24, 33, 8, 43, 62, 3, 71, 45, 48, 6, 99, 82, 76, 60, 80, 90, 68, 51, 27, 18, 56, 63, 74, 1, 61, 42, 41, 4, 15, 17, 40, 38, 5, 91, 59, 0, 34, 28, 50, 11, 35, 23, 52, 10, 31, 66, 57, 79, 85, 32, 84, 14, 89, 19, 29, 49, 97, 98, 69, 20, 94, 72, 77, 25, 37, 81, 46, 39, 65, 58, 12, 88, 70;
            \item OOD: 9, 21, 36, 44, 47, 64, 67, 83, 87, 96;
        \end{itemize}
        
        \item Split 2
        \begin{itemize}
            \item ID: 80, 84, 33, 81, 93, 17, 36, 82, 69, 65, 92, 39, 56, 52, 51, 32, 31, 44, 78, 10, 2, 73, 97, 62, 19, 35, 94, 27, 46, 38, 67, 99, 54, 95, 88, 40, 48, 59, 23, 34, 86, 53, 77, 15, 83, 41, 45, 91, 26, 98, 43, 55, 24, 4, 58, 49, 21, 87, 3, 74, 30, 66, 70, 42, 47, 89, 8, 60, 0, 90, 57, 22, 61, 63, 7, 96, 13, 68, 85, 14, 29, 28, 11, 18, 20, 50, 25, 6, 71, 76;
            \item OOD: 1, 5, 9, 12, 16, 37, 64, 72, 75, 79;
        \end{itemize}
        
        \item Split 3
        \begin{itemize}
            \item ID: 83, 30, 56, 24, 16, 23, 2, 27, 28, 13, 99, 92, 76, 14, 0, 21, 3, 29, 61, 79, 35, 11, 84, 44, 73, 5, 25, 77, 74, 62, 65, 1, 18, 48, 36, 78, 6, 89, 91, 10, 12, 53, 87, 54, 95, 32, 19, 26, 60, 55, 9, 96, 17, 59, 57, 41, 64, 45, 97, 8, 71, 94, 90, 98, 86, 80, 50, 52, 66, 88, 70, 46, 68, 69, 81, 58, 33, 38, 51, 42, 4, 67, 39, 37, 20, 31, 63, 47, 85, 93;
            \item OOD: 7, 15, 22, 34, 40, 43, 49, 72, 75, 82;
        \end{itemize}
    \end{itemize}
    
    \item \car{}
    \begin{itemize}
        \item Split 1 
        \begin{itemize}
            \item ID: 83, 12, 33, 113, 171, 134, 163, 124, 74, 18, 7, 5, 125, 161, 170, 181, 123, 60, 44, 141, 56, 173, 136, 89, 63, 55, 110, 166, 175, 45, 22, 155, 66, 37, 4, 80, 178, 106, 160, 26, 139, 143, 71, 8, 61, 130, 122, 101, 118, 92, 185, 24, 30, 109, 40, 137, 150, 90, 19, 149, 180, 54, 159, 16, 51, 177, 135, 108, 86, 96, 183, 146, 193, 14, 27, 97, 156, 46, 191, 190, 62, 2, 59, 111, 112, 43, 10, 98, 73, 188, 179, 144, 129, 93, 107, 154, 187, 50, 0, 94, 145, 95, 64, 138, 41, 69, 49, 48, 85, 13, 164, 23, 182, 131, 20, 15, 78, 104, 52, 100, 76, 3, 116, 194, 126, 6, 68, 75, 84, 121, 157, 167, 152, 91, 168, 11, 119, 102, 35, 57, 65, 1, 120, 158, 42, 105, 132, 169, 17, 38;
            \item OOD: 9, 21, 25, 28, 29, 31, 32, 34, 36, 39, 47, 53, 58, 67, 70, 72, 77, 79, 81, 82, 87, 88, 99, 103, 114, 115, 117, 127, 128, 133, 140, 142, 147, 148, 151, 153, 162, 165, 172, 174, 176, 184, 186, 189, 192, 195;
        \end{itemize}
        
        \item Split 2
        \begin{itemize}
            \item ID: 127, 11, 110, 124, 18, 165, 44, 28, 171, 136, 51, 29, 105, 56, 53, 174, 16, 164, 143, 99, 27, 130, 78, 168, 108, 191, 31, 35, 173, 4, 113, 112, 116, 82, 186, 69, 169, 40, 102, 182, 58, 14, 47, 19, 81, 39, 95, 118, 73, 67, 59, 33, 176, 154, 181, 195, 138, 117, 161, 90, 42, 17, 5, 159, 66, 48, 54, 190, 98, 89, 163, 107, 12, 120, 119, 114, 151, 84, 145, 123, 13, 188, 175, 45, 93, 36, 158, 183, 150, 97, 94, 193, 103, 170, 75, 21, 91, 155, 2, 70, 85, 147, 6, 106, 0, 152, 77, 65, 55, 122, 88, 179, 46, 62, 74, 92, 189, 184, 194, 87, 180, 177, 132, 10, 34, 32, 162, 38, 83, 153, 100, 125, 23, 126, 9, 167, 104, 148, 135, 111, 185, 64, 15, 41, 160, 109, 80, 52, 26, 76;
            \item OOD: 1, 3, 7, 8, 20, 22, 24, 25, 30, 37, 43, 49, 50, 57, 60, 61, 63, 68, 71, 72, 79, 86, 96, 101, 115, 121, 128, 129, 131, 133, 134, 137, 139, 140, 141, 142, 144, 146, 149, 156, 157, 166, 172, 178, 187, 192;
        \end{itemize}
        
        \item Split 3
        \begin{itemize}
            \item ID: 10, 79, 165, 143, 163, 134, 35, 137, 25, 2, 12, 128, 180, 3, 48, 29, 14, 119, 6, 23, 108, 144, 130, 175, 45, 120, 174, 125, 9, 164, 54, 13, 109, 169, 78, 114, 44, 82, 159, 123, 161, 94, 57, 41, 66, 98, 5, 53, 111, 11, 183, 24, 147, 64, 8, 195, 171, 28, 131, 71, 42, 65, 135, 113, 173, 189, 106, 140, 155, 187, 126, 92, 1, 85, 188, 146, 156, 184, 89, 36, 176, 100, 62, 0, 27, 153, 172, 178, 20, 154, 152, 139, 77, 30, 150, 17, 59, 186, 112, 166, 127, 158, 93, 115, 21, 55, 16, 142, 91, 99, 118, 193, 74, 122, 157, 141, 192, 87, 90, 84, 18, 97, 101, 160, 133, 61, 81, 138, 68, 129, 56, 19, 86, 70, 60, 34, 40, 76, 149, 26, 32, 191, 96, 83, 110, 105, 73, 117, 145, 151;
            \item OOD: 4, 7, 15, 22, 31, 33, 37, 38, 39, 43, 46, 47, 49, 50, 51, 52, 58, 63, 67, 69, 72, 75, 80, 88, 95, 102, 103, 104, 107, 116, 121, 124, 132, 136, 148, 162, 167, 168, 170, 177, 179, 181, 182, 185, 190, 194;
        \end{itemize}
    \end{itemize}
    
    \item \butterfly{}
    \begin{itemize}
        \item Split 1 
        \begin{itemize}
            \item ID: 018, 170, 107, 098, 177, 182, 005, 146, 012, 152, 061, 125, 180, 154, 080, 007, 033, 130, 037, 074, 183, 145, 045, 159, 060, 123, 179, 185, 122, 044, 016, 055, 150, 111, 022, 189, 129, 004, 083, 106, 134, 066, 026, 113, 168, 063, 008, 075, 118, 143, 071, 124, 184, 097, 149, 024, 030, 160, 040, 056, 131, 096, 181, 019, 153, 092, 054, 163, 051, 086, 139, 090, 137, 101, 144, 089, 109, 014, 027, 141, 187, 046, 138, 195, 108, 062, 002, 059, 136, 197, 043, 010, 194, 073, 196, 178, 175, 126, 093, 112, 158, 191, 050, 094, 110, 095, 064, 167, 041, 069, 049, 048, 085, 013, 161, 023, 186, 135, 020, 015, 078, 104, 052, 100, 076, 003, 116, 164, 198, 006, 068, 084, 121, 155, 171, 156, 091, 199, 011, 119, 102, 035, 057, 065, 001, 120, 162, 042, 105;
            \item OOD: 009, 017, 021, 025, 028, 029, 031, 032, 034, 036, 038, 039, 047, 053, 058, 067, 070, 072, 077, 079, 081, 082, 087, 088, 099, 103, 114, 115, 117, 127, 128, 132, 133, 140, 142, 147, 148, 151, 157, 165, 166, 169, 172, 173, 174, 176, 188, 190, 192, 193;
        \end{itemize}
        
        \item Split 2
        \begin{itemize}
            \item ID: 058, 040, 034, 102, 184, 198, 095, 004, 029, 168, 171, 018, 011, 089, 110, 118, 159, 035, 136, 059, 051, 016, 044, 094, 031, 162, 038, 028, 193, 027, 047, 165, 194, 177, 176, 097, 174, 073, 069, 172, 108, 107, 189, 014, 056, 019, 114, 039, 185, 124, 098, 123, 119, 053, 033, 179, 181, 106, 199, 138, 116, 067, 078, 042, 017, 005, 127, 105, 048, 066, 054, 084, 183, 158, 166, 113, 012, 117, 093, 120, 154, 090, 081, 122, 191, 013, 082, 132, 187, 045, 099, 036, 161, 186, 153, 103, 195, 197, 148, 173, 075, 021, 091, 152, 002, 070, 085, 150, 006, 112, 155, 077, 065, 055, 167, 088, 130, 046, 062, 074, 092, 147, 160, 143, 087, 180, 145, 164, 010, 032, 083, 182, 100, 125, 023, 126, 009, 170, 104, 151, 135, 111, 188, 064, 015, 041, 163, 109, 080;
            \item OOD: 001, 003, 007, 008, 020, 022, 024, 025, 026, 030, 037, 043, 049, 050, 052, 057, 060, 061, 063, 068, 071, 072, 076, 079, 086, 096, 101, 115, 121, 128, 129, 131, 133, 134, 137, 139, 140, 141, 142, 144, 146, 149, 156, 157, 169, 175, 178, 190, 192, 196;
        \end{itemize}
        
        \item Split 3
        \begin{itemize}
            \item ID: 112, 029, 182, 199, 193, 085, 010, 054, 115, 035, 012, 092, 013, 126, 174, 002, 044, 003, 113, 014, 023, 025, 006, 134, 165, 173, 045, 065, 048, 122, 178, 064, 009, 057, 078, 071, 128, 176, 131, 053, 137, 163, 111, 123, 109, 141, 041, 130, 140, 005, 159, 100, 011, 187, 024, 089, 066, 008, 172, 175, 028, 133, 094, 042, 169, 082, 184, 106, 108, 143, 180, 166, 146, 079, 001, 119, 192, 149, 160, 188, 147, 036, 171, 179, 062, 027, 157, 098, 118, 020, 158, 156, 142, 077, 030, 154, 017, 059, 181, 114, 127, 139, 191, 093, 151, 021, 055, 016, 152, 091, 099, 120, 197, 074, 190, 161, 144, 196, 087, 090, 084, 018, 097, 101, 125, 164, 135, 061, 081, 068, 129, 056, 019, 086, 070, 060, 034, 040, 138, 076, 153, 026, 032, 195, 096, 083, 110, 105, 073;
            \item OOD: 004, 007, 015, 022, 031, 033, 037, 038, 039, 043, 046, 047, 049, 050, 051, 052, 058, 063, 067, 069, 072, 075, 080, 088, 095, 102, 103, 104, 107, 116, 117, 121, 124, 132, 136, 145, 148, 150, 155, 162, 167, 168, 170, 177, 183, 185, 186, 189, 194, 198;
        \end{itemize}
    \end{itemize}
    
    \item \bird{}
    \begin{itemize}
        \item Split 1 
        \begin{itemize}
            \item ID: 0518, 0400, 0530, 0461, 0510, 0560, 0478, 0626, 0371, 0495, 0509, 0631, 0469, 0549, 0513, 0381, 0604, 0531, 0517, 0457, 0548, 0353, 0638, 0498, 0470, 0651, 0316, 0363, 0618, 0345, 0556, 0486, 0471, 0558, 0528, 0505, 0547, 0313, 0330, 0551, 0485, 0320, 0546, 0539, 0450, 0489, 0619, 0352, 0372, 0542, 0379, 0522, 0519, 0393, 0367, 0612, 0628, 0326, 0620, 0323, 0512, 0382, 0315, 0449, 0544, 0645, 0341, 0468, 0516, 0625, 0642, 0611, 0452, 0324, 0456, 0392, 0467, 0536, 0299, 0476, 0369, 0484, 0360, 0374, 0334, 0630, 0514, 0623, 0348, 0321, 0298, 0466, 0332, 0338, 0624, 0368, 0364, 0499, 0482, 0327, 0550, 0496, 0362, 0601, 0397, 0359, 0297, 0616, 0464, 0494, 0446, 0318, 0504, 0541, 0634, 0349, 0322, 0335, 0358, 0632, 0328, 0540, 0354, 0533, 0647, 0483, 0609, 0538, 0370, 0607, 0526, 0490, 0515, 0497, 0351, 0559, 0506, 0521, 0472, 0603, 0458, 0520, 0453, 0561, 0640, 0473, 0295, 0454, 0455, 0377, 0357, 0356, 0402, 0649, 0501, 0331, 0621, 0503, 0395, 0460, 0491, 0602, 0650, 0314, 0376, 0401, 0481, 0635, 0615, 0637, 0643, 0554, 0451, 0614, 0629, 0319, 0479, 0462, 0343, 0365, 0529, 0373, 0296, 0480, 0600, 0350, 0465, 0492, 0617, 0325, 0346, 0493, 0361, 0524, 0646, 0488, 0342, 0336, 0543, 0474;
            \item OOD: 0317, 0329, 0333, 0337, 0339, 0340, 0344, 0347, 0355, 0366, 0375, 0378, 0380, 0394, 0396, 0398, 0399, 0447, 0448, 0459, 0463, 0475, 0477, 0487, 0500, 0502, 0507, 0508, 0511, 0523, 0525, 0527, 0532, 0534, 0535, 0537, 0545, 0552, 0553, 0555, 0557, 0599, 0605, 0606, 0608, 0610, 0613, 0622, 0627, 0633, 0636, 0639, 0641, 0644, 0648;
        \end{itemize}
        
        \item Split 2
        \begin{itemize}
            \item ID: 0542, 0342, 0470, 0472, 0623, 0467, 0519, 0335, 0642, 0546, 0651, 0370, 0299, 0625, 0352, 0529, 0492, 0359, 0487, 0498, 0381, 0477, 0548, 0528, 0626, 0448, 0346, 0505, 0462, 0490, 0375, 0320, 0453, 0395, 0402, 0459, 0347, 0551, 0644, 0608, 0512, 0607, 0479, 0339, 0326, 0616, 0527, 0327, 0534, 0366, 0341, 0533, 0343, 0319, 0337, 0295, 0518, 0544, 0641, 0510, 0541, 0629, 0633, 0336, 0324, 0618, 0648, 0355, 0603, 0348, 0450, 0649, 0377, 0474, 0480, 0550, 0635, 0605, 0540, 0367, 0451, 0650, 0628, 0523, 0455, 0482, 0322, 0521, 0530, 0329, 0449, 0364, 0615, 0363, 0476, 0401, 0361, 0637, 0539, 0466, 0627, 0350, 0325, 0313, 0547, 0483, 0356, 0478, 0640, 0374, 0378, 0620, 0646, 0537, 0543, 0362, 0645, 0458, 0630, 0520, 0468, 0561, 0457, 0549, 0399, 0557, 0465, 0610, 0454, 0604, 0398, 0484, 0524, 0532, 0321, 0621, 0531, 0353, 0507, 0463, 0344, 0602, 0638, 0555, 0508, 0392, 0514, 0617, 0554, 0297, 0643, 0314, 0611, 0599, 0394, 0473, 0373, 0511, 0354, 0382, 0452, 0545, 0447, 0503, 0536, 0318, 0522, 0340, 0496, 0526, 0400, 0559, 0525, 0515, 0460, 0485, 0331, 0486, 0317, 0614, 0464, 0513, 0553, 0495, 0471, 0636, 0372, 0323, 0349, 0631, 0469, 0397, 0360, 0334, 0393, 0351, 0332, 0556, 0298, 0613;
            \item OOD: 0296, 0315, 0316, 0328, 0330, 0333, 0338, 0345, 0357, 0358, 0365, 0368, 0369, 0371, 0376, 0379, 0380, 0396, 0446, 0456, 0461, 0475, 0481, 0488, 0489, 0491, 0493, 0494, 0497, 0499, 0500, 0501, 0502, 0504, 0506, 0509, 0516, 0517, 0535, 0538, 0552, 0558, 0560, 0600, 0601, 0606, 0609, 0612, 0619, 0622, 0624, 0632, 0634, 0639, 0647;
        \end{itemize}
        
        \item Split 3
        \begin{itemize}
            \item ID: 0361, 0321, 0504, 0454, 0519, 0379, 0402, 0343, 0298, 0478, 0550, 0362, 0451, 0394, 0373, 0318, 0643, 0349, 0382, 0636, 0625, 0543, 0453, 0469, 0626, 0609, 0488, 0401, 0332, 0338, 0372, 0611, 0487, 0328, 0336, 0541, 0491, 0450, 0486, 0461, 0620, 0619, 0331, 0621, 0627, 0322, 0337, 0531, 0326, 0529, 0557, 0640, 0333, 0475, 0480, 0542, 0511, 0500, 0521, 0624, 0617, 0374, 0352, 0490, 0314, 0320, 0515, 0313, 0603, 0297, 0616, 0525, 0473, 0553, 0471, 0533, 0524, 0493, 0459, 0523, 0449, 0628, 0317, 0547, 0648, 0353, 0472, 0560, 0647, 0502, 0356, 0350, 0507, 0539, 0645, 0365, 0629, 0513, 0396, 0474, 0605, 0642, 0630, 0514, 0633, 0482, 0296, 0319, 0614, 0535, 0623, 0316, 0526, 0622, 0512, 0548, 0501, 0494, 0466, 0518, 0458, 0618, 0516, 0329, 0520, 0479, 0551, 0637, 0499, 0395, 0552, 0612, 0468, 0540, 0599, 0532, 0399, 0495, 0602, 0549, 0638, 0517, 0344, 0452, 0370, 0295, 0335, 0460, 0632, 0600, 0561, 0544, 0503, 0325, 0367, 0536, 0641, 0509, 0483, 0363, 0324, 0604, 0546, 0606, 0447, 0506, 0457, 0485, 0369, 0398, 0651, 0376, 0489, 0364, 0327, 0446, 0378, 0497, 0368, 0342, 0348, 0498, 0393, 0613, 0334, 0340, 0646, 0456, 0400, 0470, 0465, 0545, 0381, 0477, 0510, 0505, 0559, 0650, 0538, 0496;
            \item OOD: 0299, 0315, 0323, 0330, 0339, 0341, 0345, 0346, 0347, 0351, 0354, 0355, 0357, 0358, 0359, 0360, 0366, 0371, 0375, 0377, 0380, 0392, 0397, 0448, 0455, 0462, 0463, 0464, 0467, 0476, 0481, 0484, 0492, 0508, 0522, 0527, 0528, 0530, 0534, 0537, 0554, 0555, 0556, 0558, 0601, 0607, 0608, 0610, 0615, 0631, 0634, 0635, 0639, 0644, 0649;
        \end{itemize}
    \end{itemize}
\end{itemize}

\section{Hyperparameters of the methods}
\label{ap:hyperparam}

In Tab. \ref{tab:hyperparam} we list the candidate values and final determined values for each method's hyperparameters.
We make a few notes on the meaning of each symbol.
For ODIN and Energy, $\tau$ is the temperature factor used to scale the model logits.
Note, we do not use input pre-processing for ODIN as it is known that the perturbation hyperparameter may need exact test OOD distribution to tune \cite{hendrycks2018deep,hsu2020generalized}, and ODIN's performance is largely attributed to the temperature scaling rather than the pre-processing
\cite{roady2020open}.
For Rotation, OE, and OE-M, $\beta$ is the weighting term that controls the contribution of their regularization objective.
For EnergyOE, $m_{\text{in}}$ and $m_{\text{out}}$ is the energy threshold for ID and training outlier data, respectively.
We use $\beta=0.1$ for EnergyOE, as we find that a larger $\beta$ can hurt ID accuracy.
For Mix and MixOE training, $\beta$ again weights the regularization strength (see Eqn. \ref{eq:mixoe}).
$\alpha$ is used to parameterize the Beta distribution for the mixing operations.

\begin{table}[!ht]
\centering
\caption{The candidate values and final determined values for the hyperparameters of each method.}
\resizebox{.95\textwidth}{!}{
\begin{tabular}{lllllllll}
\toprule

& \multicolumn{2}{c}{Aircraft}
& \multicolumn{2}{c}{Car}
& \multicolumn{2}{c}{Butterfly}
& \multicolumn{2}{c}{Bird}
\\ \cmidrule(lr){2-3} \cmidrule(lr){4-5} \cmidrule(lr){6-7} \cmidrule(lr){8-9}

& Candidates &Determined
& Candidates &Determined
& Candidates &Determined
& Candidates &Determined
\\\midrule

ODIN \cite{liang2018enhancing}
&$\tau=\{10, 100, 1000\}$ &$\tau=1000$ 
&$\tau=\{10, 100, 1000\}$ &$\tau=1000$ 
&$\tau=\{10, 100, 1000\}$ &$\tau=1000$ 
&$\tau=\{10, 100, 1000\}$ &$\tau=1000$ 
\\

Energy \cite{liu2020energy}
&$\tau=\{1, 10, 100\}$ &$\tau=1$ 
&$\tau=\{1, 10, 100\}$ &$\tau=1$ 
&$\tau=\{1, 10, 100\}$ &$\tau=1$ 
&$\tau=\{1, 10, 100\}$ &$\tau=1$ 
\\

Rotation \cite{ahmed2020detecting}
&$\beta=\{0.5,1.0\}$ &$\beta=1.0$
&$\beta=\{0.5,1.0\}$ &$\beta=1.0$
&$\beta=\{0.5,1.0\}$ &$\beta=1.0$
&$\beta=\{0.5,1.0\}$ &$\beta=1.0$
\\


OE \cite{hendrycks2018deep}
&$\beta=\{0.5,1.0,5.0\}$ &$\beta=1.0$
&$\beta=\{0.5,1.0,5.0\}$ &$\beta=1.0$
&$\beta=\{0.5,1.0,5.0\}$ &$\beta=1.0$
&$\beta=\{0.5,1.0,5.0\}$ &$\beta=1.0$
\\

OE-M \cite{chen2021atom}
&$\beta=\{0.5,1.0,5.0\}$ &$\beta=1.0$
&$\beta=\{0.5,1.0,5.0\}$ &$\beta=1.0$
&$\beta=\{0.5,1.0,5.0\}$ &$\beta=1.0$
&$\beta=\{0.5,1.0,5.0\}$ &$\beta=1.0$
\\ [1mm] \cdashline{1-9} \\[-3mm]

EnergyOE \cite{liu2020energy}
&$m_{\text{in}}=\{-9,-13,-17\}$ &$m_{\text{in}}=-17$

&$m_{\text{in}}=\{-9,-13,-17\}$ &$m_{\text{in}}=-13$

&$m_{\text{in}}=\{-10,-14,-18\}$ &$m_{\text{in}}=-14$

&$m_{\text{in}}=\{-9,-13,-17\}$ &$m_{\text{in}}=-13$
\\

&$m_{\text{out}}=\{-5,-7,-9\}$ 
&$m_{\text{out}}=-9$

&$m_{\text{out}}=\{-4,-6,-8\}$ 
&$m_{\text{out}}=-6$

&$m_{\text{out}}=\{-5,-7,-9\}$ 
&$m_{\text{out}}=-5$

&$m_{\text{out}}=\{-5,-7,-9\}$ 
&$m_{\text{out}}=-5$
\\ [1mm] \cdashline{1-9} \\[-3mm]

Mix-\textit{linear} \cite{zhang2018mixup}
&$\alpha=\{0.4,1.0,2.0\}$
&$\alpha=1.0$

&$\alpha=\{0.4,1.0,2.0\}$
&$\alpha=2.0$

&$\alpha=\{0.4,1.0,2.0\}$
&$\alpha=1.0$

&$\alpha=\{0.4,1.0,2.0\}$
&$\alpha=1.0$
\\

&$\beta=\{0.5,1.0,5.0\}$
&$\beta=1.0$

&$\beta=\{0.5,1.0,5.0\}$
&$\beta=5.0$

&$\beta=\{0.5,1.0,5.0\}$
&$\beta=0.5$

&$\beta=\{0.5,1.0,5.0\}$
&$\beta=0.5$
\\ [1mm] \cdashline{1-9} \\[-3mm]

Mix-\textit{cut} \cite{yun2019cutmix}
&$\alpha=\{0.4,1.0,2.0\}$
&$\alpha=2.0$

&$\alpha=\{0.4,1.0,2.0\}$
&$\alpha=2.0$

&$\alpha=\{0.4,1.0,2.0\}$
&$\alpha=0.4$

&$\alpha=\{0.4,1.0,2.0\}$
&$\alpha=0.4$
\\

&$\beta=\{0.5,1.0,5.0\}$
&$\beta=5.0$

&$\beta=\{0.5,1.0,5.0\}$
&$\beta=5.0$

&$\beta=\{0.5,1.0,5.0\}$
&$\beta=0.5$

&$\beta=\{0.5,1.0,5.0\}$
&$\beta=0.5$
\\ [1mm] \cdashline{1-9} \\[-3mm]

MixOE-\textit{linear} 
&$\alpha=\{0.4,1.0,2.0\}$
&$\alpha=1.0$

&$\alpha=\{0.4,1.0,2.0\}$
&$\alpha=1.0$

&$\alpha=\{0.4,1.0,2.0\}$
&$\alpha=1.0$

&$\alpha=\{0.4,1.0,2.0\}$
&$\alpha=1.0$
\\

&$\beta=\{0.5,1.0,5.0\}$
&$\beta=5.0$

&$\beta=\{0.5,1.0,5.0\}$
&$\beta=5.0$

&$\beta=\{0.5,1.0,5.0\}$
&$\beta=5.0$

&$\beta=\{0.5,1.0,5.0\}$
&$\beta=5.0$
\\ [1mm] \cdashline{1-9} \\[-3mm]

MixOE-\textit{cut}
&$\alpha=\{0.4,1.0,2.0\}$
&$\alpha=2.0$

&$\alpha=\{0.4,1.0,2.0\}$
&$\alpha=2.0$

&$\alpha=\{0.4,1.0,2.0\}$
&$\alpha=1.0$

&$\alpha=\{0.4,1.0,2.0\}$
&$\alpha=1.0$
\\

&$\beta=\{0.5,1.0,5.0\}$
&$\beta=5.0$

&$\beta=\{0.5,1.0,5.0\}$
&$\beta=5.0$

&$\beta=\{0.5,1.0,5.0\}$
&$\beta=5.0$

&$\beta=\{0.5,1.0,5.0\}$
&$\beta=5.0$
\\

\bottomrule

\end{tabular}
}
\label{tab:hyperparam}
\end{table}

\section{Visualization procedure}
\label{ap:visualize_detail}
Following \cite{pang2019rethinking}, the visualization is enabled by training a ``visualization'' layer (denoted as vis layer for brevity).
Specifically, the vis layer is a fully-connected layer that maps the output of DNN's penultimate layer (in our case with ResNet-50 the output is a 2048-dimensional vector) to a 2D space.
When training such vis layer, we concatenate it to the penultimate layer of the ResNet and add another consequent linear layer that maps the 2D output of the vis layer to the class logits.
Critically, the vis layer is trained by minimizing the cross-entropy loss on ID training data only; no OOD data are ever involved.
This ensures that the process remains agnostic to any OOD samples.
Also, during the training the preceding layers of the DNN model are freezed, so the visualization is faithful to the model.
After the training, the visualization is performed by simply passing inputs into the model and taking the 2D outputs of the vis layer.

\section{Analysis on ablation study}
\label{ap:ablation_study_analysis}

In this section, we conduct further analysis to explain the ablation study results presented in Fig. \ref{fig:ablation}.

\begin{figure}[!t]
  \centering
  \includegraphics[width=.95\linewidth]{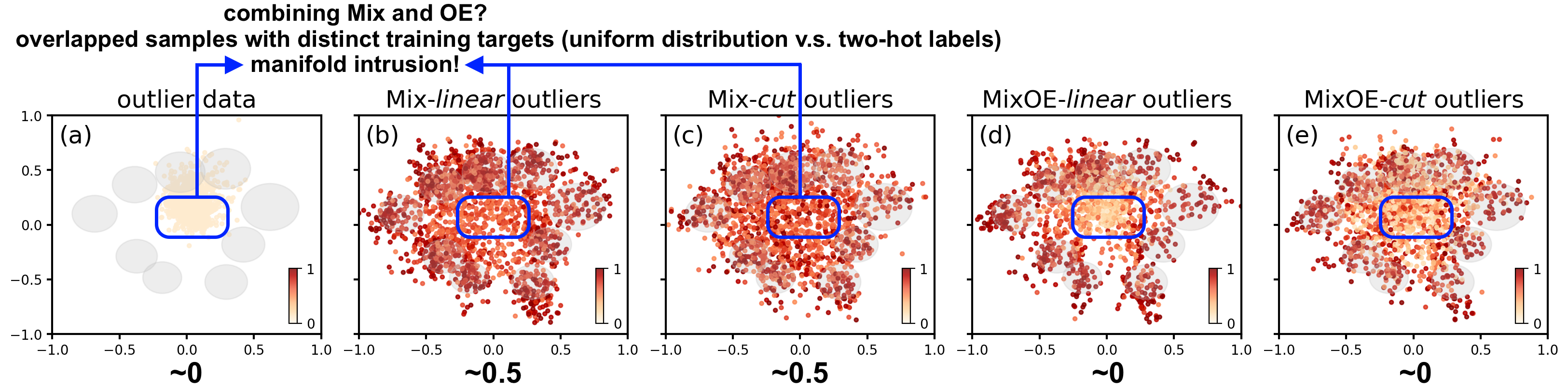}
  \caption{2D Visualization of the data representations in the DNN's feature space.
  The color lightness indicates the confidence encoded in the soft target of the corresponding outlier sample. The numbers below the figures are the prediction confidence each method assigns to the outliers within the {\color{blue} blue region} during training.
  Compared with Mix training, our MixOE will enforce a more desirable confidence decay (from the {\color{high} highest level} to the {\color{low} lowest level}) as the inputs transitions from ID to OOD. 
  This explains why MixOE outperforms Mix.
  Also, blindly combining Mix and OE will \textit{not} work because of manifold intrusion \cite{guo2019mixup}.
  See the text for detailed analysis.}
  \label{fig:method_vis_ablation}
\end{figure}

\noindent \textbf{Why MixOE is better than Mix training.}
To explain these results, using the same idea as Fig. \ref{fig:method_vis}, in Fig. \ref{fig:method_vis_ablation} we visualize the training outliers used by Mix and MixOE in the feature space of a standardly trained model (see Fig. \ref{fig:vis_all_aircraft}-\ref{fig:vis_all_bird} for more).

Similar to MixOE, Mix training's data can also cover a broader space of the OOD region, which accounts for their improved performance against fine-grained OOD data.
However, one problem we notice is that the confidence encoded in the soft targets of Mix training is still relatively high (above 50\%) even when the corresponding inputs are already far away from the ID clusters.
This is because when interpolating between two one-hot labels (as Mix does), the minimum confidence that can be achieved is 50\%.
Consequently, during Mix training the prediction confidence of the OOD samples that are relatively far away from the ID clusters may not be pushed towards the lowest level (\ie, resembling uniform distribution as in MixOE; see Fig. \ref{fig:method_vis_ablation}). 
Such consequence can be confirmed by the distribution plots of Mix training's prediction confidence in Fig. \ref{fig:conf_all_aircraft}-\ref{fig:conf_all_bird}.
This reasoning explains why Mix training underperforms MixOE (especially its little effect on coarse OOD data) and further validates MixOE's idea of using ID and outlier data together to regularize the prediction confidence over the OOD region.

\noindent \textbf{MixOE v.s. Mix + OE.}
Upon the previous discussion, one may wonder if naively combining the Mix and OE objective together can achieve similar effect to MixOE.
Yet, a crucial problem with such combined objective is that it will assign distinct training targets to two overlapped groups of samples.
Specifically, from the blue regions in Fig. \ref{fig:method_vis_ablation} (a)/(b)/(c), one can see that the auxiliary outlier data can overlap with the mixed samples generated by the Mix training.
However, their corresponding soft targets are very different: mixed samples are coupled with ``two-hot'' targets that result from the interpolation between two one-hot labels, yet the auxiliary outlier data are trained to match uniform distribution.
This inconsistency is similar to the problem of manifold intrusion \cite{guo2019mixup} and may cause learning difficulty for the model.
Indeed, like we stated in the main text, when combining Mix and OE together, the model's accuracy can decrease by up to 10\%, and the TNR95 can be worse than MSP by 10\% and 20\% against coarse- and fine-grained OOD samples, respectively.

\section{Additional plots and results}
\label{ap:additional_results}

\begin{figure*}[!t]
  \centering
  \includegraphics[width=0.8\linewidth]{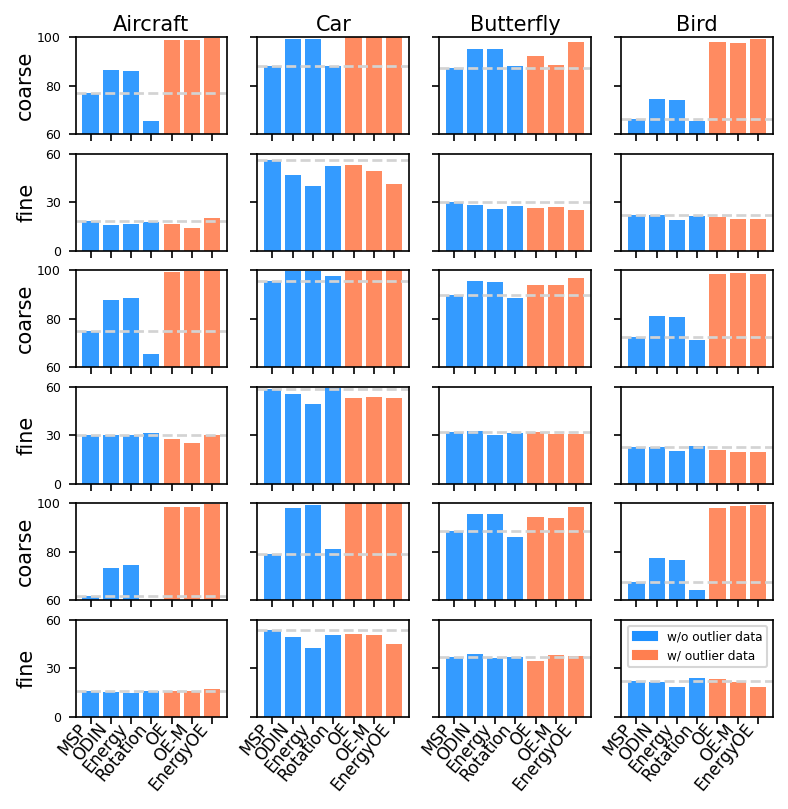}
  \caption{TNR95 statistics of existing methods on all of the three splits in each dataset. Each two rows correspond to one split.}
  \label{fig:initial_eval_all}
\end{figure*}

\begin{figure*}[!t]
  \centering
  \includegraphics[width=0.87\linewidth]{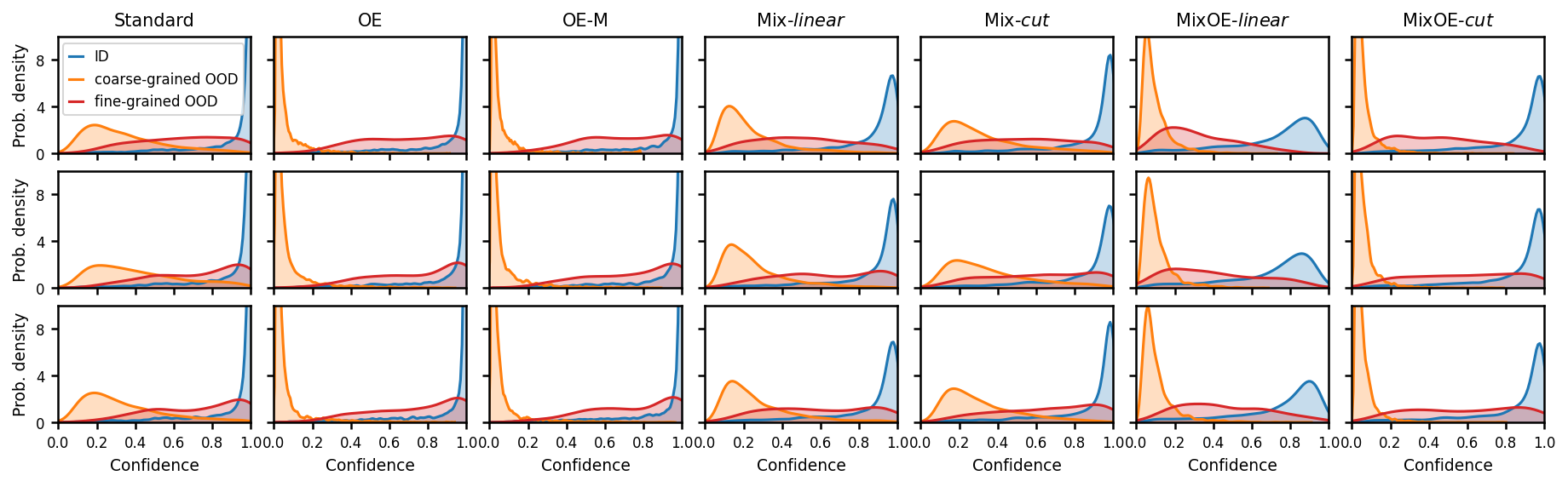}
  \vspace{-2mm}
  \caption{Confidence density plots on the three splits in the \aircraft{} dataset.}
  \label{fig:conf_all_aircraft}
  \vspace{-3mm}
\end{figure*}

\begin{figure*}[!h]
  \centering
  \includegraphics[width=0.87\linewidth]{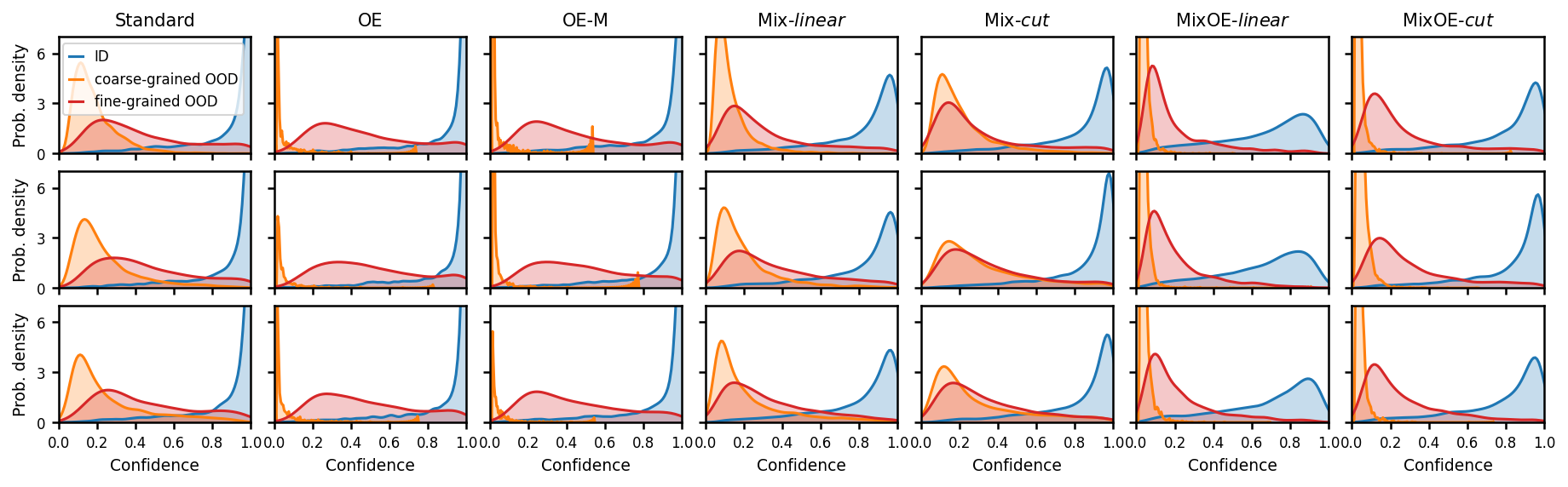}
  \vspace{-2mm}
  \caption{Confidence density plots on the three splits in the \car{} dataset.}
  \label{fig:conf_all_car}
  \vspace{-3mm}
\end{figure*}

\begin{figure*}[!h]
  \centering
  \includegraphics[width=0.87\linewidth]{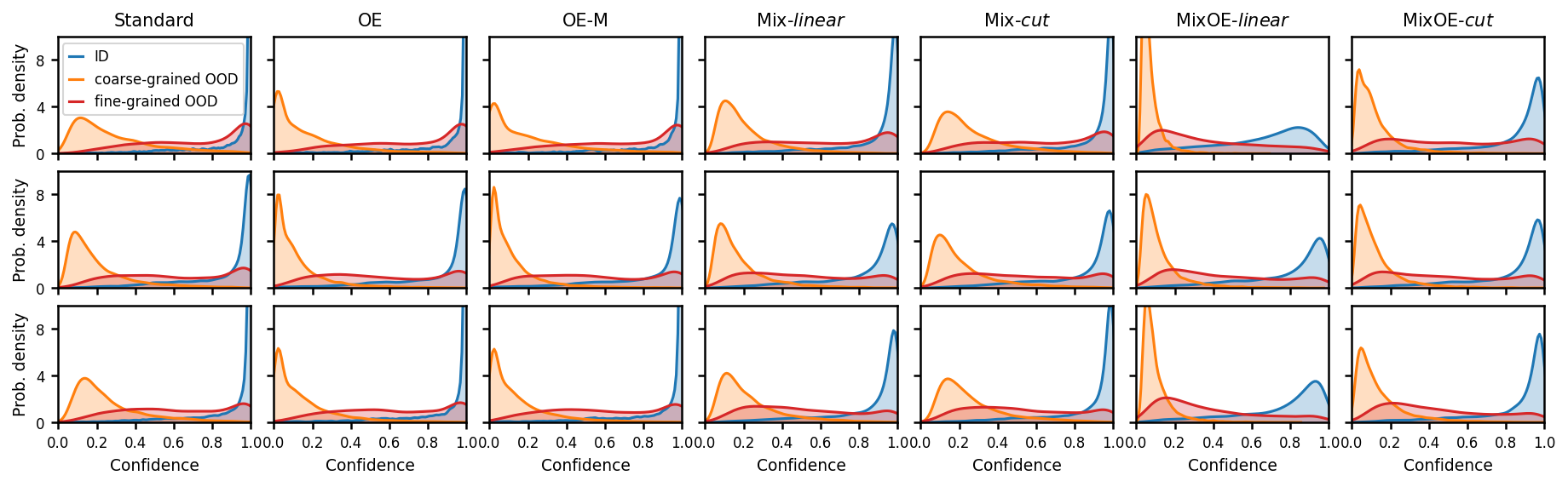}
  \vspace{-2mm}
  \caption{Confidence density plots on the three splits in the \butterfly{} dataset.}
  \label{fig:conf_all_butterfly}
  \vspace{-3mm}
\end{figure*}

\begin{figure*}[!h]
  \centering
  \includegraphics[width=0.87\linewidth]{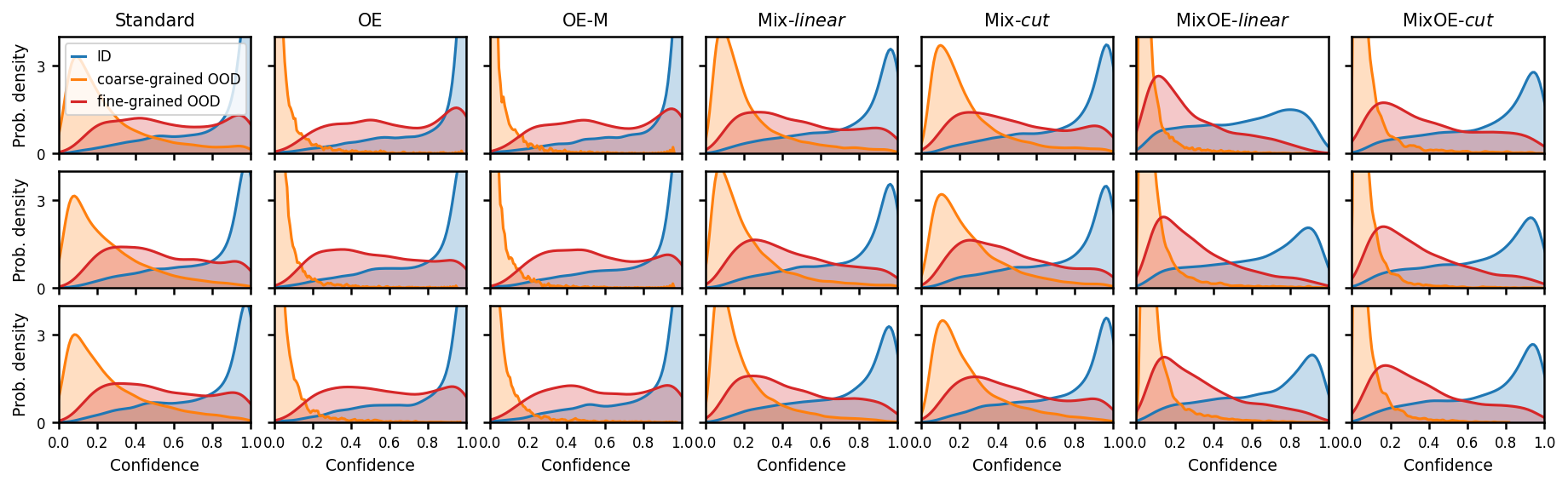}
  \vspace{-2mm}
  \caption{Confidence density plots on the three splits in the \bird{} dataset.}
  \label{fig:conf_all_bird}
  \vspace{-3mm}
\end{figure*}

\begin{figure*}[!t]
  \centering
  \includegraphics[width=0.9\linewidth]{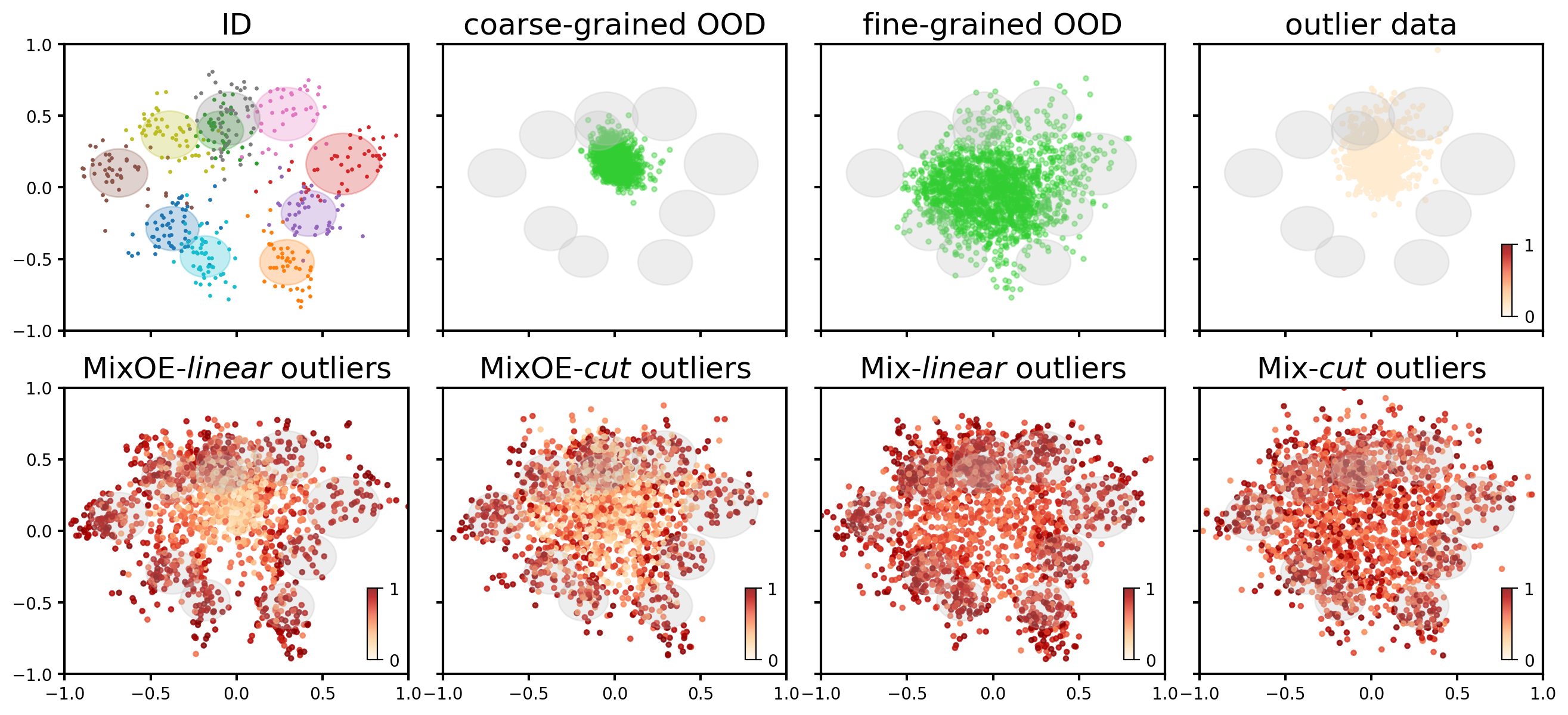}
  \caption{2D visualization of ID/OOD/training outlier data in the feature space for the \aircraft{} dataset.}
  \label{fig:vis_all_aircraft}
\end{figure*}

\begin{figure*}[!t]
  \centering
  \includegraphics[width=0.9\linewidth]{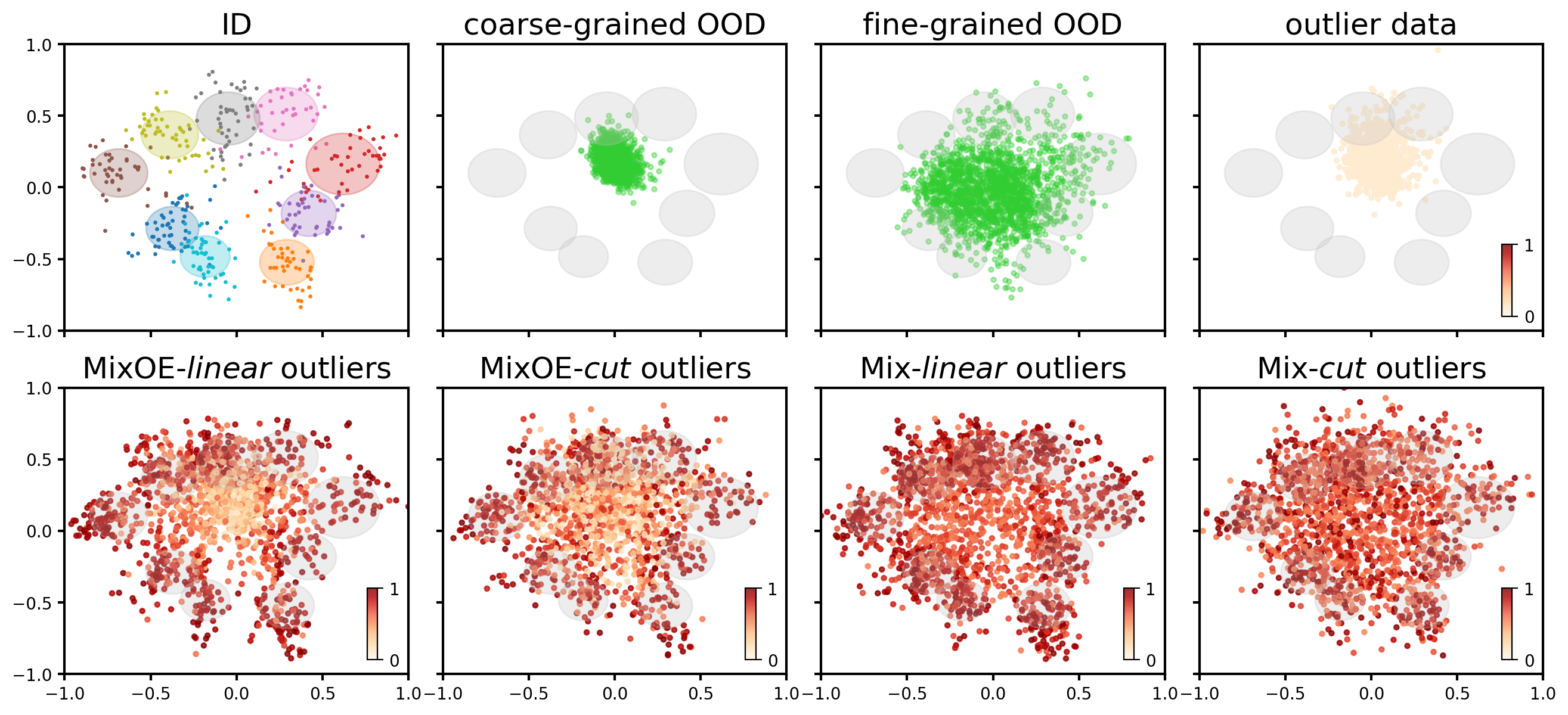}
  \caption{2D visualization of ID/OOD/training outlier data in the feature space for the \car{} dataset.}
  \label{fig:vis_all_car}
\end{figure*}

\begin{figure*}[!t]
  \centering
  \includegraphics[width=0.9\linewidth]{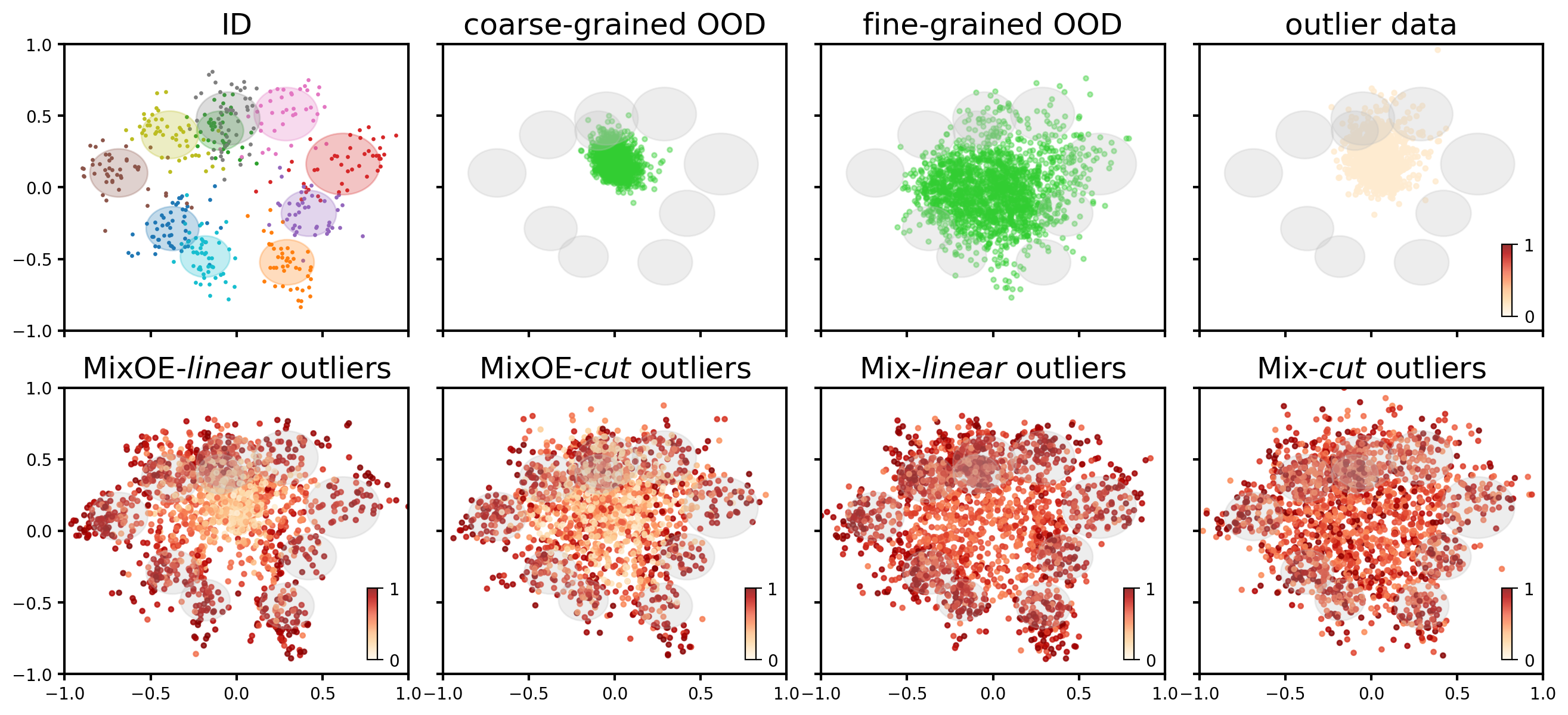}
  \caption{2D visualization of ID/OOD/training outlier data in the feature space for the \butterfly{} dataset.}
  \label{fig:vis_all_butterfly}
\end{figure*}

\begin{figure*}[!t]
  \centering
  \includegraphics[width=0.9\linewidth]{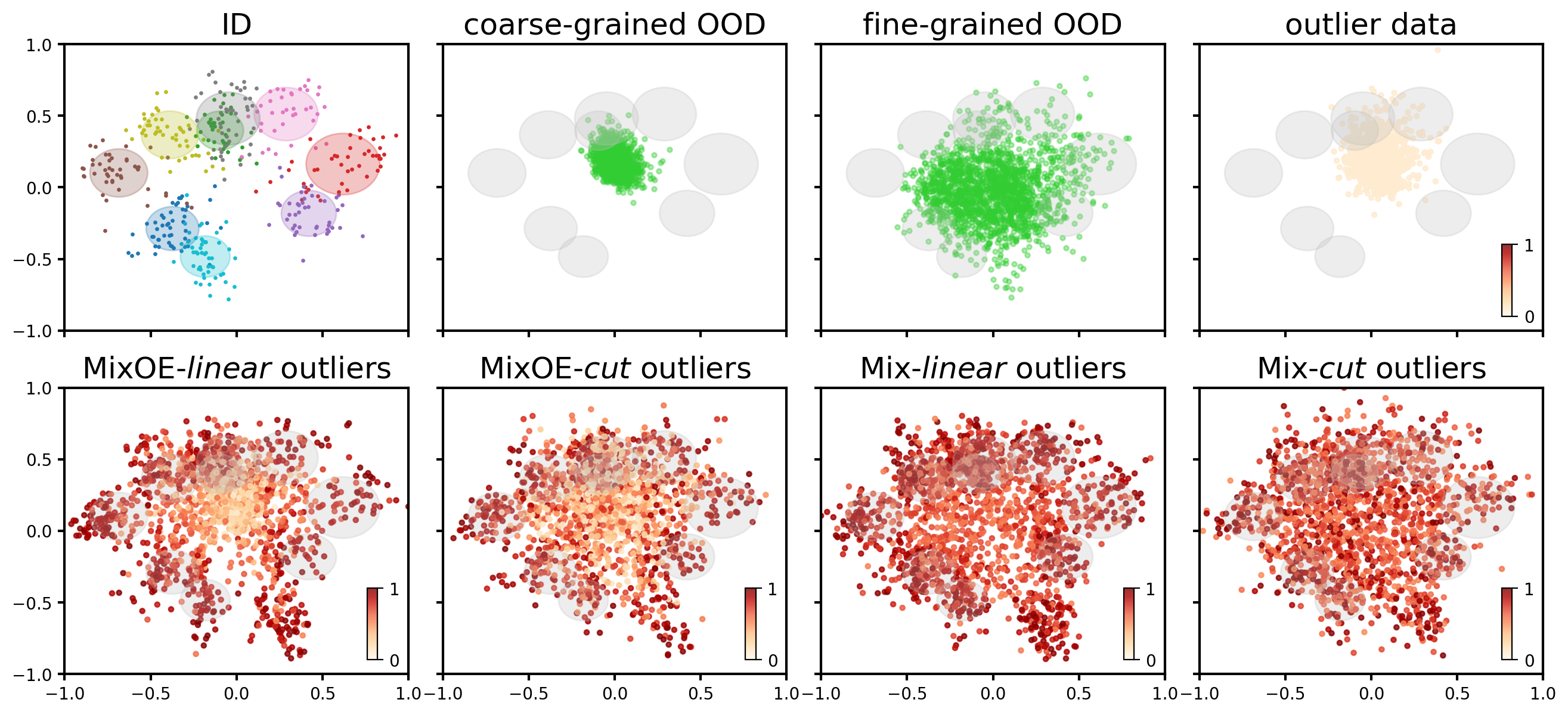}
  \caption{2D visualization of ID/OOD/training outlier data in the feature space for the \bird{} dataset.}
  \label{fig:vis_all_bird}
\end{figure*}

Here we provide additional plots and results that are not shown in the main document due to space limit.
Tab. \ref{tab:full_results} presents the full detection results in terms of both TNR95 and AUROC for all methods on all splits/datasets.
Fig. \ref{fig:initial_eval_all} includes the bar plots that visualize the TNR95 statistics of existing methods on all splits/datasets.
The corresponding numerical results can be found in Tab. \ref{tab:full_results}.
Fig. \ref{fig:conf_all_aircraft}-\ref{fig:conf_all_bird} show the confidence distributions of the Standard, OE, Corruption, Mix, and MixOE models on all splits/datasets.
Fig. \ref{fig:vis_all_aircraft}-\ref{fig:vis_all_bird} show the 2D visualization of ID/OOD/training outlier data on all datasets.
Tab. \ref{tab:accuracy_full_results} presents the accuracy results of all methods on all splits/datasets.

\begin{table}[!ht]
\centering
\caption{Full detection results. The number before and after the slash is for coarse-grained and fine-grained OOD samples, respectively. Avg. diff. is the average difference (across three splits) relative to MSP.} 
\resizebox{0.95\textwidth}{!}{
\begin{tabular}{llrrrrrrrr}
\toprule
\multirow{2}{*}{$\mathcal{D}_{\text{in}}$} & \multirow{2}{*}{Method} & \multicolumn{2}{c}{Split 1} &\multicolumn{2}{c}{Split 2} &\multicolumn{2}{c}{Split 3} &\multicolumn{2}{c}{Avg. diff.} \\ \cmidrule(lr){3-4} \cmidrule(lr){5-6} \cmidrule(lr){7-8} \cmidrule(lr){9-10}
& & \multicolumn{1}{r}{TNR95} & \multicolumn{1}{r}{AUROC} & \multicolumn{1}{r}{TNR95} & \multicolumn{1}{r}{AUROC} & \multicolumn{1}{r}{TNR95} & \multicolumn{1}{r}{AUROC} & \multicolumn{1}{r}{TNR95} & \multicolumn{1}{r}{AUROC}\\ \midrule

\multirow{11}{*}{\rotatebox[origin=c]{90}{Aircraft}}
&MSP \cite{hendrycks2017baseline} 
&75.0 / 29.9  &95.6 / 85.9  &61.6 / 15.9  &93.6 / 78.7  &77.1 / 18.5  &95.8 / 77.6  &- / -  &- / -
\\

&ODIN \cite{liang2018enhancing}
&87.5 / 30.2  &97.4 / 85.5  &73.2 / 15.3  &95.6 / 75.5  &86.5 / 15.8  &97.2 / 78.6  &\textcolor{\upcolor}{+11.2} / \textcolor{\downcolor}{--1.0}  &\textcolor{\upcolor}{+1.7} / \textcolor{\downcolor}{--0.9}
\\

&Energy \cite{liu2020energy} 
&88.5 / 30.1  &97.5 / 85.2  &74.4 / 14.6  &95.6 / 74.9  &86.2 / 16.3  &97.2 / 78.3  &\textcolor{\upcolor}{+11.8} / \textcolor{\downcolor}{--1.1}  &\textcolor{\upcolor}{+1.8} / \textcolor{\downcolor}{--1.3}
\\

&Rotation \cite{ahmed2020detecting}
&65.5 / 31.4  &94.0 / 86.0  &55.0 / 15.9  &92.2 / 79.1  &65.5 / 17.6  &93.4 / 77.6  &\textcolor{\downcolor}{--9.2} / \textcolor{\upcolor}{+0.2}  &\textcolor{\downcolor}{--1.8} / \textcolor{\upcolor}{+0.2}
\\

&Mix-\textit{linear} \cite{zhang2018mixup}
&80.6 / 34.2  &96.2 / 86.6  &80.5 / 19.8  &96.6 / 77.6  &76.8 / 20.9  &95.6 / 78.4  &\textcolor{\upcolor}{+8.1} / \textcolor{\upcolor}{+3.5}  &\textcolor{\upcolor}{+1.1} / \textcolor{\upcolor}{+0.1}
\\

&Mix-\textit{cut} \cite{yun2019cutmix}
&68.1 / 33.6  &94.3 / 86.6  &63.1 / 24.3  &92.9 / 78.0  &70.1 / 16.9  &94.9 / 77.4  &\textcolor{\downcolor}{--4.1} / \textcolor{\upcolor}{+3.5}  &\textcolor{\downcolor}{--1.0} / \textcolor{\downcolor}{--0.1}
\\[1mm] \cdashline{2-10} \\[-2mm]



&OE \cite{hendrycks2018deep}
&99.3 / 27.8  &99.8 / 84.6  &98.5 / 16.0  &99.6 / 78.4  &98.7 / 16.5  &99.7 / 78.7  &\textcolor{\upcolor}{+27.6} / \textcolor{\downcolor}{--1.3}  &\textcolor{\upcolor}{+4.7} / \textcolor{\downcolor}{--0.2}
\\

&OE-M \cite{chen2021atom}
&99.6 / 25.0  &99.9 / 85.1  &98.5 / 16.0  &99.7 / 79.2  &98.9 / 14.0  &99.7 / 78.1  &\textcolor{\upcolor}{+27.8} / \textcolor{\downcolor}{--3.1}  &\textcolor{\upcolor}{+4.8} / \textcolor{\upcolor}{+0.1}
\\

&EnergyOE \cite{liu2020energy} 
&99.8 / 30.3  &99.9 / 86.9  &99.7 / 17.0  &99.8 / 76.7  &99.7 / 19.9  &99.8 / 79.1  &\textcolor{\upcolor}{+28.5} / \textcolor{\upcolor}{+1.0}  &\textcolor{\upcolor}{+4.8} / \textcolor{\upcolor}{+0.2}
\\ 

\rowcolor{gray!15}\cellcolor{white} &\method-\textit{linear}
&93.2 / 41.4  &98.3 / 89.5  &88.4 / 24.6  &97.9 / 81.8  &92.1 / 16.5  &98.3 / 80.9  &\textcolor{\upcolor}{+20.0} / \textcolor{\upcolor}{+6.1}  &\textcolor{\upcolor}{+3.2} / \textcolor{\upcolor}{+3.3}
\\

\rowcolor{gray!15}\cellcolor{white} &\method-\textit{cut}
&99.0 / 39.8  &99.7 / 89.4  &99.4 / 23.7  &99.7 / 80.1  &99.4 / 24.9  &99.8 / 78.6  &\textcolor{\upcolor}{+28.0} / \textcolor{\upcolor}{+8.0}  &\textcolor{\upcolor}{+4.7} / \textcolor{\upcolor}{+2.0}
\\
\midrule

\multirow{11}{*}{\rotatebox[origin=c]{90}{Car}}
&MSP \cite{hendrycks2017baseline} 
&95.5 / 58.5  &98.8 / 90.2  &88.0 / 56.3  &97.8 / 90.6  &78.8 / 53.5  &96.4 / 90.2  &- / -  &- / -
\\

&ODIN \cite{liang2018enhancing}
&99.6 / 55.6  &99.7 / 89.0  &99.1 / 47.0  &99.5 / 88.2  &97.8 / 49.0  &99.2 / 88.7  &\textcolor{\upcolor}{+11.4} / \textcolor{\downcolor}{--5.6}  &\textcolor{\upcolor}{+1.8} / \textcolor{\downcolor}{--1.7}
\\

&Energy \cite{liu2020energy} 
&99.7 / 49.1  &99.8 / 88.1  &99.4 / 39.7  &99.7 / 86.8  &99.1 / 42.6  &99.6 / 87.7  &\textcolor{\upcolor}{+12.0} / \textcolor{\downcolor}{--12.3}  &\textcolor{\upcolor}{+2.0} / \textcolor{\downcolor}{--2.8}
\\

&Rotation \cite{ahmed2020detecting}
&97.7 / 58.9  &99.3 / 90.5  &88.1 / 52.4  &97.6 / 90.1  &81.3 / 50.4  &96.9 / 90.0  &\textcolor{\upcolor}{+1.6} / \textcolor{\downcolor}{--2.2}  &\textcolor{\upcolor}{+0.3} / \textcolor{\downcolor}{--0.1}
\\



&Mix-\textit{linear} \cite{zhang2018mixup}
&95.9 / 66.3  &98.8 / 91.9  &82.6 / 52.3  &96.7 / 89.0  &77.8 / 58.6  &95.5 / 91.2  &\textcolor{\downcolor}{--2.0} / \textcolor{\upcolor}{+3.0}  &\textcolor{\downcolor}{--0.7} / \textcolor{\upcolor}{+0.4}
\\

&Mix-\textit{cut} \cite{yun2019cutmix}
&84.6 / 68.8  &97.0 / 92.0  &70.6 / 65.3  &93.8 / 92.2  &70.6 / 61.6  &93.1 / 91.6  &\textcolor{\downcolor}{--12.2} / \textcolor{\upcolor}{+9.1}  &\textcolor{\downcolor}{--3.0} / \textcolor{\upcolor}{+1.6}
\\[1mm] \cdashline{2-10} \\[-2mm]

&OE \cite{hendrycks2018deep}
&99.9 / 53.2  &100.0 / 89.9  &100.0 / 53.0  &100.0 / 89.7  &99.9 / 51.2  &100.0 / 90.2  &\textcolor{\upcolor}{+12.5} / \textcolor{\downcolor}{--3.6}  &\textcolor{\upcolor}{+2.3} / \textcolor{\downcolor}{--0.4}
\\

&OE-M \cite{chen2021atom}
&99.9 / 53.6  &100.0 / 88.9  &100.0 / 49.4  &100.0 / 89.5  &100.0 / 50.6  &100.0 / 90.1  &\textcolor{\upcolor}{+12.5} / \textcolor{\downcolor}{--4.9}  &\textcolor{\upcolor}{+2.3} / \textcolor{\downcolor}{--0.8}
\\

&EnergyOE \cite{liu2020energy} 
&100.0 / 52.6  &100.0 / 89.0  &100.0 / 41.0  &100.0 / 87.2  &100.0 / 44.9  &100.0 / 88.3  &\textcolor{\upcolor}{+12.6} / \textcolor{\downcolor}{--9.9}  &\textcolor{\upcolor}{+2.3} / \textcolor{\downcolor}{--2.2}
\\ 

\rowcolor{gray!15}\cellcolor{white} &\method-\textit{linear}
&99.6 / 65.9  &99.8 / 92.2  &99.7 / 62.9  &99.8 / 92.3  &99.5 / 60.1  &99.8 / 92.0  &\textcolor{\upcolor}{+12.2} / \textcolor{\upcolor}{+6.9}  &\textcolor{\upcolor}{+2.1} / \textcolor{\upcolor}{+1.8}
\\

\rowcolor{gray!15}\cellcolor{white} &\method-\textit{cut}
&99.9 / 70.3  &99.9 / 92.6  &100.0 / 69.8  &99.9 / 93.0  &99.9 / 66.5  &99.9 / 93.1  &\textcolor{\upcolor}{+12.5} / \textcolor{\upcolor}{+12.8}  &\textcolor{\upcolor}{+2.2} / \textcolor{\upcolor}{+2.6}
\\
\midrule

\multirow{11}{*}{\rotatebox[origin=c]{90}{Butterfly}}
&MSP \cite{hendrycks2017baseline} 
&87.1 / 29.9  &97.8 / 79.3  &89.9 / 31.8  &97.9 / 78.2  &88.4 / 36.6  &97.9 / 83.8  &- / -  &- / -
\\

&ODIN \cite{liang2018enhancing}
&95.2 / 28.2  &98.7 / 74.5  &95.5 / 32.5  &98.8 / 77.3  &95.6 / 38.7  &98.8 / 82.5  &\textcolor{\upcolor}{+7.0} / \textcolor{\upcolor}{+0.4}  &\textcolor{\upcolor}{+0.9} / \textcolor{\downcolor}{--2.3}
\\

&Energy \cite{liu2020energy} 
&95.3 / 25.5  &98.7 / 73.9  &95.2 / 30.2  &98.7 / 76.7  &95.6 / 36.1  &98.7 / 82.0  &\textcolor{\upcolor}{+6.9} / \textcolor{\downcolor}{--2.2}  &\textcolor{\upcolor}{+0.8} / \textcolor{\downcolor}{--2.9}
\\

&Rotation \cite{ahmed2020detecting}
&87.9 / 27.6  &97.8 / 78.0  &88.5 / 31.2  &97.9 / 79.7  &86.2 / 37.0  &97.5 / 83.9  &\textcolor{\downcolor}{--0.9} / \textcolor{\downcolor}{--0.8}  &\textcolor{\downcolor}{--0.1} / \textcolor{\upcolor}{+0.1}
\\

&Mix-\textit{linear} \cite{zhang2018mixup}
&90.7 / 31.3  &98.1 / 78.5  &88.7 / 34.9  &97.8 / 80.6  &84.7 / 37.9  &97.2 / 83.6  &\textcolor{\downcolor}{--0.4} / \textcolor{\upcolor}{+1.9}  &\textcolor{\downcolor}{--0.2} / \textcolor{\upcolor}{+0.5}
\\

&Mix-\textit{cut} \cite{yun2019cutmix}
&85.7 / 30.4  &97.3 / 78.6  &86.3 / 34.6  &97.2 / 79.7  &84.7 / 40.5  &97.1 / 84.3  &\textcolor{\downcolor}{--2.9} / \textcolor{\upcolor}{+2.4}  &\textcolor{\downcolor}{--0.7} / \textcolor{\upcolor}{+0.4}
\\[1mm] \cdashline{2-10} \\[-2mm]



&OE \cite{hendrycks2018deep}
&92.2 / 26.5  &98.4 / 77.8  &93.7 / 32.1  &98.4 / 79.3  &94.3 / 34.3  &98.6 / 82.6  &\textcolor{\upcolor}{+4.9} / \textcolor{\downcolor}{--1.8}  &\textcolor{\upcolor}{+0.6} / \textcolor{\downcolor}{--0.5}
\\

&OE-M \cite{chen2021atom}
&88.5 / 26.9  &97.7 / 78.0  &94.0 / 30.5  &98.4 / 79.0  &93.9 / 37.8  &98.4 / 83.6  &\textcolor{\upcolor}{+3.7} / \textcolor{\downcolor}{--1.0}  &\textcolor{\upcolor}{+0.3} / \textcolor{\downcolor}{--0.2}
\\

&EnergyOE \cite{liu2020energy} 
&97.8 / 25.1  &99.1 / 73.5  &96.9 / 30.5  &98.9 / 75.9  &98.2 / 37.2  &99.1 / 82.6  &\textcolor{\upcolor}{+9.2} / \textcolor{\downcolor}{--1.8}  &\textcolor{\upcolor}{+1.2} / \textcolor{\downcolor}{--3.1}
\\ 

\rowcolor{gray!15}\cellcolor{white} &\method-\textit{linear}
&95.3 / 32.6  &98.7 / 78.3  &93.9 / 37.9  &98.6 / 80.9  &95.5 / 45.0  &98.8 / 84.9  &\textcolor{\upcolor}{+6.4} / \textcolor{\upcolor}{+5.7}  &\textcolor{\upcolor}{+0.8} / \textcolor{\upcolor}{+0.9}
\\

\rowcolor{gray!15}\cellcolor{white} &\method-\textit{cut}
&94.9 / 35.8  &98.8 / 79.8  &94.1 / 38.8  &98.7 / 81.7  &92.7 / 46.0  &98.6 / 86.8  &\textcolor{\upcolor}{+5.4} / \textcolor{\upcolor}{+7.4}  &\textcolor{\upcolor}{+0.8} / \textcolor{\upcolor}{+2.3}
\\
\midrule

\multirow{11}{*}{\rotatebox[origin=c]{90}{Bird}}
&MSP \cite{hendrycks2017baseline} 
&72.3 / 22.6  &93.7 / 76.4  &67.4 / 22.3  &93.5 / 78.0  &66.4 / 22.3  &92.9 / 76.9  &- / -  &- / -
\\

&ODIN \cite{liang2018enhancing}
&80.9 / 22.7  &95.3 / 75.5  &77.2 / 21.5  &95.5 / 76.6  &74.3 / 21.9  &94.5 / 75.8  &\textcolor{\upcolor}{+8.8} / \textcolor{\downcolor}{--0.4}  &\textcolor{\upcolor}{+1.7} / \textcolor{\downcolor}{--1.1}
\\

&Energy \cite{liu2020energy} 
&80.8 / 20.3  &95.3 / 74.8  &76.5 / 18.4  &95.4 / 75.5  &73.9 / 18.8  &94.4 / 74.8  &\textcolor{\upcolor}{+8.4} / \textcolor{\downcolor}{--3.2}  &\textcolor{\upcolor}{+1.7} / \textcolor{\downcolor}{--2.1}
\\

&Rotation \cite{ahmed2020detecting}
&71.3 / 23.6  &93.9 / 76.0  &64.0 / 24.0  &92.7 / 78.7  &65.4 / 21.5  &92.8 / 76.8  &\textcolor{\downcolor}{--1.8} / \textcolor{\upcolor}{+0.6}  &\textcolor{\downcolor}{--0.2} / \textcolor{\upcolor}{+0.1}
\\

&Mix-\textit{linear} \cite{zhang2018mixup}
&72.1 / 24.4  &94.0 / 77.6  &70.0 / 25.3  &94.0 / 79.9  &71.0 / 25.0  &94.3 / 79.5  &\textcolor{\upcolor}{+2.3} / \textcolor{\upcolor}{+2.5}  &\textcolor{\upcolor}{+0.7} / \textcolor{\upcolor}{+1.9}
\\

&Mix-\textit{cut} \cite{yun2019cutmix}
&69.0 / 25.6  &93.5 / 77.1  &58.7 / 23.6  &92.1 / 79.2  &65.3 / 25.0  &92.9 / 77.9  &\textcolor{\downcolor}{--4.4} / \textcolor{\upcolor}{+2.3}  &\textcolor{\downcolor}{--0.5} / \textcolor{\upcolor}{+1.0}
\\[1mm] \cdashline{2-10} \\[-2mm]


&OE \cite{hendrycks2018deep}
&98.2 / 20.6  &99.5 / 75.9  &97.9 / 22.9  &99.5 / 78.2  &97.9 / 20.7  &99.5 / 77.0  &\textcolor{\upcolor}{+29.3} / \textcolor{\downcolor}{--1.0}  &\textcolor{\upcolor}{+6.1} / \textcolor{\downcolor}{--0.1}
\\

&OE-M \cite{chen2021atom}
&98.7 / 19.8  &99.6 / 76.1  &98.7 / 21.4  &99.6 / 78.2  &97.7 / 19.2  &99.4 / 76.5  &\textcolor{\upcolor}{+29.7} / \textcolor{\downcolor}{--2.3}  &\textcolor{\upcolor}{+6.2} / \textcolor{\downcolor}{--0.2}
\\

&EnergyOE \cite{liu2020energy} 
&98.6 / 19.4  &99.6 / 74.4  &99.0 / 18.4  &99.7 / 76.2  &99.3 / 19.5  &99.7 / 77.7  &\textcolor{\upcolor}{+30.3} / \textcolor{\downcolor}{--3.3}  &\textcolor{\upcolor}{+6.3} / \textcolor{\downcolor}{--1.0}
\\ 

\rowcolor{gray!15}\cellcolor{white} &\method-\textit{linear}
&88.6 / 24.9  &97.6 / 77.8  &83.9 / 26.7  &96.6 / 80.6  &86.3 / 28.6  &97.1 / 80.7  &\textcolor{\upcolor}{+17.6} / \textcolor{\upcolor}{+4.3}  &\textcolor{\upcolor}{+3.7} / \textcolor{\upcolor}{+2.6}
\\

\rowcolor{gray!15}\cellcolor{white} &\method-\textit{cut}
&91.0 / 27.7  &98.0 / 78.6  &91.8 / 24.6  &98.4 / 80.2  &92.9 / 27.7  &98.7 / 80.0  &\textcolor{\upcolor}{+23.2} / \textcolor{\upcolor}{+4.3}  &\textcolor{\upcolor}{+5.0} / \textcolor{\upcolor}{+2.5}
\\
\bottomrule

\end{tabular}
}
\label{tab:full_results}
\end{table}

\clearpage

\begin{table*}[!ht]
\centering
\caption{Full accuracy results on each split of each dataset.}
\begin{tabular}{llrrr}
\toprule
$\mathcal{D}_{\text{in}}$ & Method & Split 1 &Split 2 &Split 3 \\ \midrule

\multirow{7}{*}{\rotatebox[origin=c]{90}{Aircraft}}
&Standard
&89.53  &89.41  &89.94
\\

&Rotation \cite{ahmed2020detecting}
&88.20  &88.31  &89.01
\\


&OE \cite{hendrycks2018deep}
&88.57  &88.87  &90.08
\\

&OE-M \cite{chen2021atom}
&89.50  &88.77  &89.54
\\

&EnergyOE \cite{liu2020energy} 
&89.27  &88.81  &89.88
\\ 

\rowcolor{gray!15}\cellcolor{white} &\method-\textit{linear}
&89.93  &90.81  &90.64
\\

\rowcolor{gray!15}\cellcolor{white} &\method-\textit{cut}
&89.77  &90.21  &90.31 
\\
\midrule

\multirow{7}{*}{\rotatebox[origin=c]{90}{Car}}
&Standard
&91.74  &92.19  &91.57
\\

&Rotation \cite{ahmed2020detecting}
&91.38  &91.64  &90.82
\\


&OE \cite{hendrycks2018deep}
&91.56  &91.75  &91.53
\\

&OE-M \cite{chen2021atom}
&90.58  &91.22  &91.40
\\

&EnergyOE \cite{liu2020energy} 
&91.61  &92.16  &91.76
\\ 

\rowcolor{gray!15}\cellcolor{white} &\method-\textit{linear}
&92.53  &93.53  &92.76 
\\

\rowcolor{gray!15}\cellcolor{white} &\method-\textit{cut}
&92.74  &93.46  &92.56
\\
\midrule

\multirow{7}{*}{\rotatebox[origin=c]{90}{Butterfly}}
&Standard
&90.21  &87.28  &89.38
\\

&Rotation \cite{ahmed2020detecting}
&89.75  &87.88  &88.88
\\


&OE \cite{hendrycks2018deep}
&89.74  &86.13  &88.41 
\\

&OE-M \cite{chen2021atom}
&89.99  &85.62  &88.86
\\

&EnergyOE \cite{liu2020energy} 
&90.21  &87.22  &88.86
\\ 

\rowcolor{gray!15}\cellcolor{white} &\method-\textit{linear}
&90.71  &88.04  &89.10
\\

\rowcolor{gray!15}\cellcolor{white} &\method-\textit{cut}
&91.33  &88.46  &90.53
\\
\midrule

\multirow{7}{*}{\rotatebox[origin=c]{90}{Bird}}
&Standard
&83.30  &82.03  &80.96
\\

&Rotation \cite{ahmed2020detecting}
&82.47  &82.25  &81.26
\\


&OE \cite{hendrycks2018deep}
&83.32  &82.03  &81.75 
\\

&OE-M \cite{chen2021atom}
&83.35  &82.70  &81.93
\\

&EnergyOE \cite{liu2020energy} 
&82.89  &82.57  &81.36 
\\ 

\rowcolor{gray!15}\cellcolor{white} &\method-\textit{linear}
&84.18  &83.29  &82.75 
\\

\rowcolor{gray!15}\cellcolor{white} &\method-\textit{cut}
&84.37  &83.44  &82.57 
\\
\bottomrule

\end{tabular}
\label{tab:accuracy_full_results}
\end{table*}

\end{document}